\documentclass{ieeeaccess}
\usepackage{cite}
\usepackage{amsmath,amssymb,amsfonts}
\usepackage{algorithmic}
\usepackage{textcomp}
\usepackage{color,soul}
\usepackage[pdftex]{graphicx}
\usepackage{subcaption}
\usepackage{setspace}
\usepackage{float}
\usepackage{caption}
\usepackage[svgnames]{xcolor}
\definecolor{blue}{rgb}{0.0, 0.53, 0.74}
\usepackage{caption}
\usepackage[labelfont={color=blue,bf}]{caption}

\usepackage{xcolor}
\usepackage{pgfplots}
\usepackage{pgfplotstable}
\usepackage{scalerel}
\pgfplotsset{compat=1.7}
\usepackage{tikz}
\usetikzlibrary{svg.path}
\NewSpotColorSpace{PANTONE}
  \AddSpotColor{PANTONE} {PANTONE3015C} {PANTONE\SpotSpace 3015\SpotSpace C} {1 0.3 0 0.2}
  \SetPageColorSpace{PANTONE}
\graphicspath{ {./images/} }
\definecolor{orcidlogocol}{HTML}{A6CE39}
\tikzset{
    orcidlogo/.pic={
        \fill[orcidlogocol] svg{M256,128c0,70.7-57.3,128-128,128C57.3,256,0,198.7,0,128C0,57.3,57.3,0,128,0C198.7,0,256,57.3,256,128z};
        \fill[white] svg{M86.3,186.2H70.9V79.1h15.4v48.4V186.2z}
        svg{M108.9,79.1h41.6c39.6,0,57,28.3,57,53.6c0,27.5-21.5,53.6-56.8,53.6h-41.8V79.1z M124.3,172.4h24.5c34.9,0,42.9-26.5,42.9-39.7c0-21.5-13.7-39.7-43.7-39.7h-23.7V172.4z}
        svg{M88.7,56.8c0,5.5-4.5,10.1-10.1,10.1c-5.6,0-10.1-4.6-10.1-10.1c0-5.6,4.5-10.1,10.1-10.1C84.2,46.7,88.7,51.3,88.7,56.8z};
    }
}

\newcommand\orcidicon[1]{\href{https://orcid.org/#1}{\mbox{\scalerel*{
                \begin{tikzpicture}[yscale=-1,transform shape]
                \pic{orcidlogo};
                \end{tikzpicture}
            }{|}}}}

\usepackage{hyperref}
\usepackage{siunitx} % Required for alignment
\usepackage{longtable}
\usepackage{tabularx}
\usepackage{placeins}
\renewcommand\hl[1]{#1}
\def\BibTeX{{\rm B\kern-.05em{\sc i\kern-.025em b}\kern-.08em
    T\kern-.1667em\lower.7ex\hbox{E}\kern-.125emX}}
\begin{document}
\history{Date of publication: 09 July 2021. Please find the latest published version by visiting the following link.}
\doi{https://doi.org/10.1109/ACCESS.2021.3096136}

\title{Recent Advances in Deep Learning Techniques for Face Recognition}
\author{\uppercase{Md. Tahmid Hasan Fuad}\authorrefmark{1},
\uppercase{Awal Ahmed Fime\authorrefmark{1}, Delowar Sikder\authorrefmark{1}, MD. Akil Raihan Iftee\authorrefmark{1}, Jakaria Rabbi$^{\textsuperscript{\orcidicon{0000-0001-9572-9010}}}$\authorrefmark{1}, Mabrook S. Al-Rakhami$^{\textsuperscript{\orcidicon{0000-0001-5343-8370}}}$\authorrefmark{2}
\IEEEmembership{Member, IEEE}, Abdu Gumae\authorrefmark{2}, Ovishake Sen\authorrefmark{1}, Mohtasim Fuad\authorrefmark{1}, and MD. NAZRUL ISLAM\authorrefmark{1}}}
\address[1]{Department of Computer Science and Engineering, Khulna University of Engineering \& Technology, Khulna-9203, Bangladesh}
\address[2]{Research Chair of Pervasive and Mobile Computing, Information Systems Department, College of Computer and Information Sciences, King Saud University, Riyadh 11543, Saudi Arabia}

\tfootnote{The authors are grateful to the Deanship of Scientific Research, king Saud University for funding through Vice Deanship of Scientific Research Chairs.}

%\markboth
%{Author \headeretal: Preparation of Papers for IEEE TRANSACTIONS and JOURNALS}
%{Author \headeretal: Preparation of Papers for IEEE TRANSACTIONS and JOURNALS}

%the professor who gives us fund will be the corresponding author as he will submit the paper. 

% one person from us will be another corresponding author.

% usually one or two person are stated as corresponding author.

%it has no significance other than "point of contact through email"

\corresp{Corresponding author: Jakaria Rabbi (jakaria\_rabbi@cse.kuet.ac.bd) and Mabrook S. Al-Rakhami (malrakhami@ksu.edu.sa)}

\begin{abstract}
In recent years, researchers have proposed many deep learning (DL) methods for various tasks, and particularly face recognition (FR) made an enormous leap using these techniques. Deep FR systems benefit from the hierarchical architecture of the DL methods to learn discriminative face representation. Therefore, DL techniques significantly improve state-of-the-art performance on FR systems and encourage diverse and efficient real-world applications. In this paper, we present a comprehensive analysis of various FR systems that leverage the different types of DL techniques, and for the study, we summarize \hl{171} recent contributions from this area.  We discuss the papers related to different algorithms, architectures, loss functions, activation functions, datasets, challenges, improvement ideas, current and future trends of DL-based FR systems. We provide a detailed discussion of various DL methods to understand the current state-of-the-art, and then we discuss various activation and loss functions for the methods. Additionally, we summarize different datasets used widely for FR tasks and discuss challenges related to illumination, expression, pose variations, and occlusion. Finally, we discuss improvement ideas, current and future trends of FR tasks.
\end{abstract}


\titlepgskip=-15pt

\maketitle

\section{Introduction}
\label{sec:introduction}
\begin{figure}[ht]
    \includegraphics[width=\linewidth]{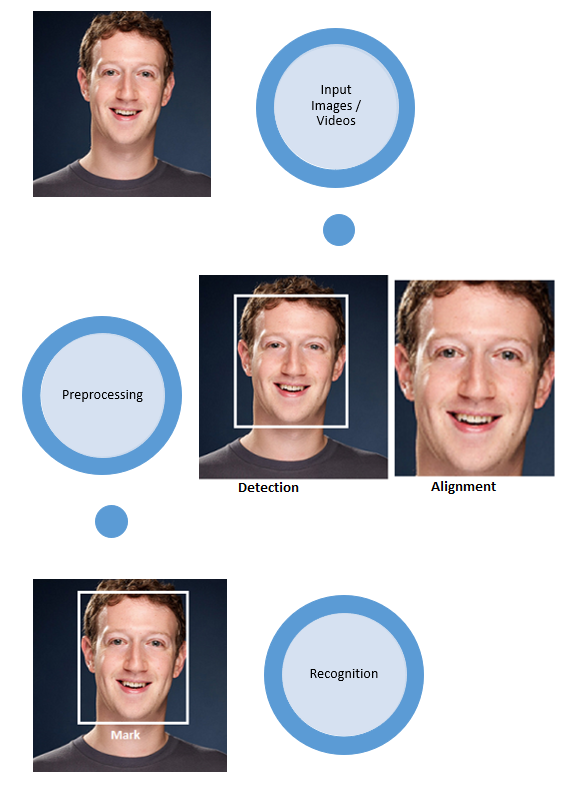}
    \caption{Face Recognition Pipeline.}
    \label{fig:face_recog_pipeline}
\end{figure}

\begin{figure}[ht]
    
    \includegraphics[width=80mm,scale=1.0]{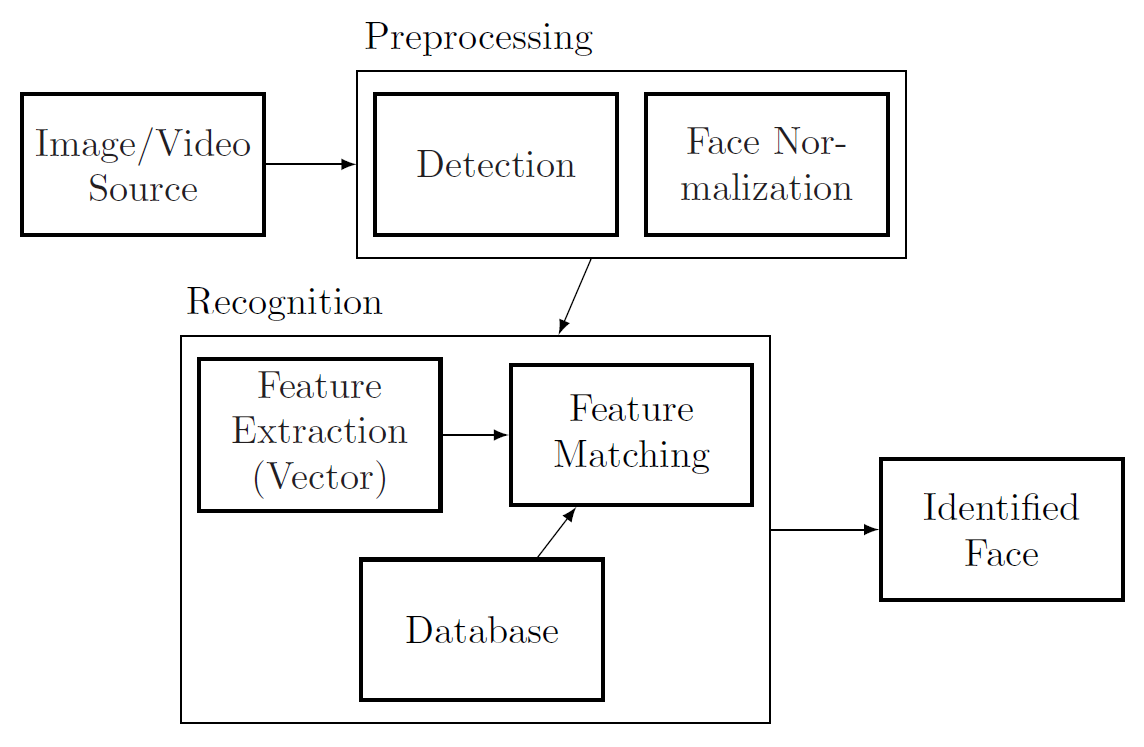}
    \caption{Face Recognition Block Diagram.}
    \label{fig:FRblock_diagram}
\end{figure}

\begin{figure}[ht]
    \centering
    \includegraphics[width=\linewidth]{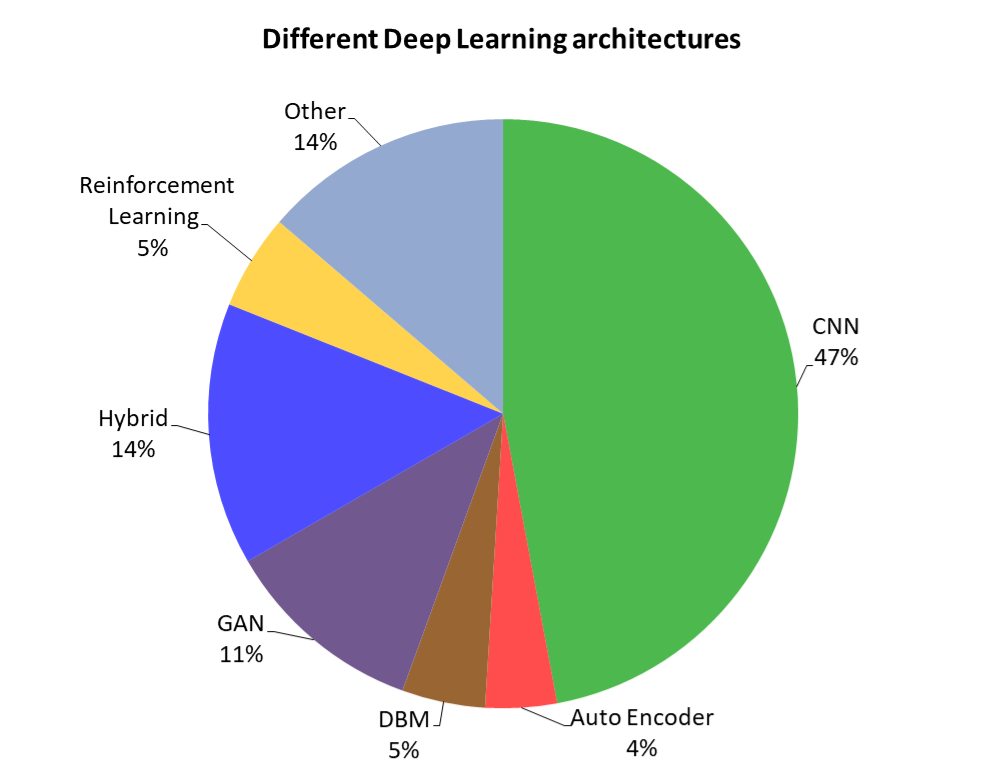}
    \caption{Different Deep Learning Architectures  for Face Recognition.}
    \label{fig:deep_learning_architecture}
\end{figure}

\begin{figure}[ht]
\begin{tikzpicture}[scale=0.9]
\begin{axis} [
xlabel={Number of Papers},
ylabel={Year},
symbolic y coords={1990 to 2012,2013,2014,2015,2016,2017,2018,2019,2020,2021},
ytick=data,
xtick={0,5,10,15,20,25,30},
xmajorgrids=true,
grid style=dashed
]
\addplot[xbar,fill=cyan] coordinates {
    (9,1990 to 2012)
    (4,2013)
	(9,2014)
	(15,2015)
	(22,2016)
	(32,2017)
	(26,2018)
	(31,2019)
	(24,2020)
	(1,2021)
};
\end{axis}
\end{tikzpicture}
\caption{Paper Distribution by Year.}
\label{fig:year-based_distribution}
\end{figure}
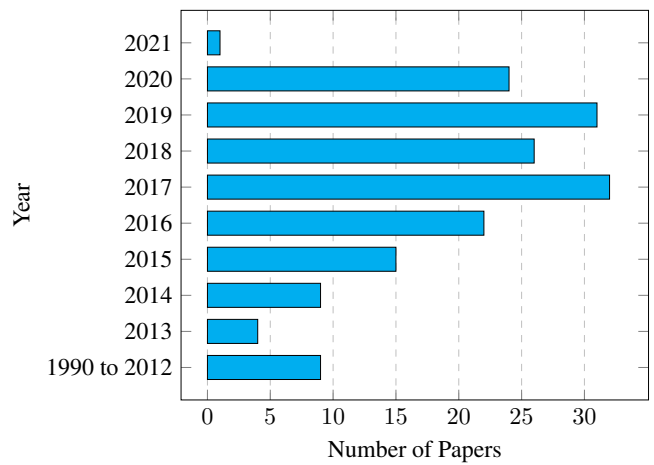

%1. general introduction 2. classical face detection 3. machine learning methods 4. ANN 5. Deep NN 6. CNN 7. Deep Reinforcement learning 8. Datasets 9. Loss function 10 Paper summary - contributions 11. organization 
%
%- Awal
%
\hl{The human face is a crucial aspect of social communication and interaction.} Humans need to recognize the face of others for these purposes. Throughout his whole life, a person has to recognize thousands of other persons' faces surrounding him. 
For human-computer interaction, face recognition is also essential. Nowadays, it is also widely used in access control, security, surveillance systems, the entertainment industry. Improvement of face recognition makes those work easier and faster. Face recognition can be divided into two types: face verification and face identification. Face verification is a 1:1 matching where it \hl{simply} detects from two images, whether both images are from the same person or not. On the other hand, face identification 1:N matching, where it is needed to determine who this person is in the image among all possible outputs.
Figure \ref{fig:face_recog_pipeline} shows us the pipeline of Face Recognition (FR) and Figure \ref{fig:FRblock_diagram} shows the block diagram of FR. FR is a combination of three sub-task\hl{s}: face detection, feature extraction or alignment, and face verification or identification. Our work mainly focus\hl{es} on feature extraction from face images and how those can be classified. Figure \ref{fig:deep_learning_architecture} shows the percentage of different face recognition techniques in our review and Figure \ref{fig:year-based_distribution} shows the year wise distribution of the papers.

\subsection{classical face detection} 
\hl{Face detection with a machine was started with some simple statistical techniques}. Eigenfaces\cite{eigenfaces} was one of the most popular among them. It represents every image as a vector of weights obtained by projecting on eigenfaces components\cite{traditional-detection-survey}. Researchers also tried to use some other traditional methods like elastic graph matching\cite{wiskott1997face}, Karhunen-Loeve based methods\cite{KL-procedure}, singular value decomposition\cite{SVD} for face recognition. Those methods were mostly tested on small datasets. Even in some cases, the size of the dataset was less than 100. Though statistical methods are not quite efficient, it gives the confidence that the machine itself can recognize the human face without external interference. It has a long-lasting certain impact on further improvement.

\subsection{feature extraction}
The performance of any model on any particular dataset largely depends on the features extracted from the data. After using various methods on face detection, face alignment is also done using various statistical methods. Some face alignment methods are Active Appearance Model (AAM)\cite{AAM}, Active Shape Model (ASM)\cite{ASM}. Those aligned faces are used for feature extraction. Some traditional methods for feature extraction are Local Binary Pattern\cite{lbp}, Fisher vectors\cite{fisher_vector}. Some dimensionality reduction methods like Principal Component Analysis (PCA)\cite{pca}, Subclass Discriminant Analysis (SDA)\cite{sda} can be used for feature extraction. Depending on the priority area for feature extraction, it can be divided into local feature extraction and global feature extraction. Global feature generalizes the whole image; for example, Histogram of Gradient (HoG) and Bag of Words (BoW) use global feature extraction.
On the other hand, local features extract the key points from the image, and one example of this method is Local Binary Patterns (LBP)\cite{lbp}. Depending on how the features are extracted, feature extraction can be divided into many types; geometry-based technique, holistic approach,  template-based technique, appearance-based approach, and color-based method are some of them. \hl{Also, several hybrid and hand-crafted\mbox{\cite{ding2015multi}} methods are traditionally used for face recognition.} From time to time, feature extraction methods have been improved and robustly extracted more advanced features. So that face recognition methods can do their job more efficiently. Nowadays, Convolutional Neural Network (CNN) or Deep Convolutional Neural Network (DCNN) based methods are primarily used for feature extraction. Taking advantage of these feature extraction methods, we can implement face recognition efficiently using traditional machine learning methods like SVM\cite{svm}.

\subsection{Artificial Neural Network}
\hl{Artificial Neural Network (ANN), a network which is inspired by the biological neurons,} got attention from various researchers. They tried to use it in face recognition in multiple forms. ANN was constructed for some specific purposes like data classification or pattern recognition. The main idea of using ANN for face recognition is to use extract features with different feature extraction methods and use them in different ANN combinations. WISARD (WIlkie, Stoneham and Aleksander's Recognition Device)\cite{wisard} was one of the initial models of ANN, which was used for face recognition\cite{survey2006}. It has a single-layer adaptive neural network structure. Taking advantage of Gabor features and LDA feature extraction models, ANN can recognize persons from images\cite{gabor-LDA-ANN}. Fernandez et al.\cite{hand-extracted-feature-ANN} started with detecting the face using Viola-Jones Algorithm and cropped it. Then they extracted the skin color, eyes color, the distance between the two eyes, the width of the nose, the height and width of the lips, and the distance between the nose and the lips from those cropped images. Those features are used in an ANN to identify a specific person.

\subsection{Deep Neural Network}
Development of Deep Neural Network (DNN) and applying them into face recognition systems push recognition further ahead. DNN can extract more diverse features effectively from inputs which are never possible for ANN or other statistical methods. DNN is an extended version of ANN with multiple hidden layers in it. With some disadvantages, the more hidden layers in the DNN network, the more robust feature it can extract. Shepley\cite{criitical_analysis_S} provided a critical analysis and comparison of different state-of-the-art DNN based face recognition methods in his survey paper and showed their benefits and problems. Learned-Miller et al.\cite{LFW_S} discussed different approaches on the LFW dataset in their work. Balaban et al. \cite{Deep_L_influence} provided a brief introduction to the influences of the state-of-the-art deep learning methods in face recognition.

\subsection{Convolutional Neural Network}
\hl{Due to} the recent progress of Deep Convolutional Neural Networks (CNNs)\cite{resnet}, \cite{vgg}, the performance of state-of-the-art methods on image processing has significantly increased. Most of the CNN-based face recognition tasks are done by following the conventional pipeline of two steps. First is face detection and then recognition of those detected faces using different network architecture \cite{PoreNet}, \cite{Millions-3D-dataset}, \cite{face-body-video}, \cite{ring_loss}. However, there are some exceptions too \cite{wu2017recursive}. CNN-based models mainly extract features from images and use them in face recognition. CNN can extract handy high-label features that are hard for a human to understand \hl{and} different studies did the extraction in different ways. FaceNet\cite{facenet} extract high-quality features from images and predict 128 elements from them and represent them in a vector named face embedding. \hl{This face embedding is} used as the basis for training classifier systems. Some researchers \cite{yang2017neural}, \cite{face-body-video} also tried to recognize face from video\hl{s}. \hl{Face tracking is the additional step to recognize a face from different frames of a video.} In some controlled environments, CNN-based face recognition can do much better than humans.

\subsection{Deep Reinforcement learning} %-- Fuad\\
Reinforcement learning (RL) comes from the eagerness to mimic humans’ decision-making process. RL agents decide their behavior from environment experience using Markov Decision Process (MDP)\cite{littman2015reinforcement}. Generally, RL is not used directly in face recognition. It is used as a part of a hybrid method like CNN and RL or GAN (Generative Adversarial Network) and RL. The researchers used RL to solve some problems, for instance, adaptation of loss functions\cite{liu2019fair}, skewness embedding\cite{wang2020mitigating}, user authentication \cite{wang2017face}, and searching a set of dominant features \cite{harandi2004face}.

\subsection{loss function}% -- Awal\\
Different deep learning-based face recognition models mainly differ in three main positions: dataset, network architecture, and loss function. A loss function is used to evaluate how well the model can predict the output by mapping with the actual output. If the model can predict the output properly, it produces a small value; otherwise, it provides a high value. Historically, with CNN\hl{,} traditional softmax\cite{sun2014deep}, \cite{chal19} can be used for face recognition. However, some studies\cite{liu2017sphereface}, \cite{center-loss} show that traditional softmax is not always quite sufficient for classification task. So, the researchers pay attention to develop more powerful loss functions. Most of the loss functions share the same idea of maximizing inter-class distance and/or minimizing intra-class distance \cite{ring_loss} \cite{center-loss}. \hl{The preference of the loss function used in a model also depends on the neural networks activation function.}

\subsection{Dataset}
Finding a properly labeled and large dataset is another important criterion for developing a new and more accurate face recognition technique. Early dataset like CASIA-WebFace \cite{casia-webface-dataset} to recent datasets like MS-Celeb-1M \cite{MS-Celeb-1M}, VggFace2 \cite{cao2018vggface2} and IMDb \cite{IMDb} are playing their role to develop new techniques. With the improvement of multimedia technology, both datasets and the number of images are increasing. CASIA-WebFace \cite{casia-webface-dataset} contains 0.5M images from 10,575 individuals. On the other hand, MSCeleb-1M \cite{MS-Celeb-1M} has about 10M images of 100K identities. As a result, there are some problems in the labeling of the data. The researchers like 

\begin{figure*}[ht]
    \includegraphics[width=\linewidth]{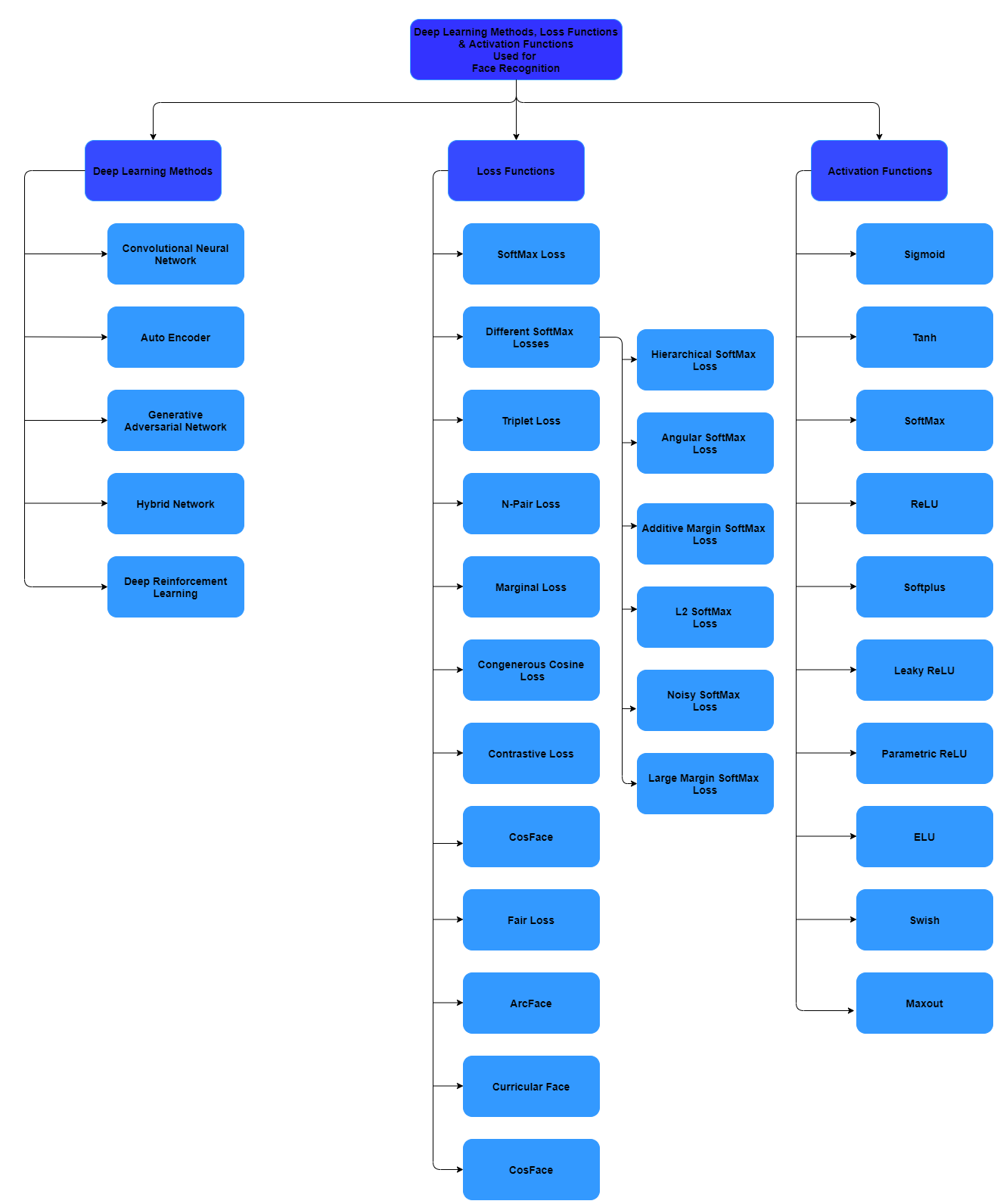}
    \caption{Taxonomy of the Deep Learning Methods, Loss Functions \& Activation Functions used for Face Recognition.}
    \label{fig:face_recog_taxonomy}
\end{figure*}
\clearpage

\hl{\noindent Wang et al. \cite{Co-mining}, Wu et al. \cite{a6}, and  Deng et al.\cite{deng2019arcface} tried to solve this problem. Depending on the labeled data, those face dataset can be divided into two types: open-set and close-set.} 

\subsection{3D Face Recognition}
3D face recognition can perform face recognition more efficiently than 2D face recognition. Because it does not face problems with light, pose, rotation, make-up, or blur images. Also, geometric information of 3D faces is more reliable than 2D faces.
Initially, LDA, PCA, color-based methods, Gaussian,  Gabor wavelet approach \cite{3D-survey-2005}, \cite{wiskott1997face} were mostly used for 3D face recognition. Most of the current methods are depending on DCNN, GAN, or pose variations. As 3D face data can not be used directly in face recognition, some pre-processing and feature extraction makes those data usable.
3D FR can be divided into two types depending on the extracted features: local feature-based methods\cite{local-feature-3D} and global \cite{global-feature-3D} feature-based methods.\hl{ There is another technique called hybrid that is also used in feature extraction.} It is a combination of local and global feature-based methods. It performs better than each technique individually.
However, the main problem with 3D face recognition is, it does not have a large dataset. It is also not possible to collect data from websites like 2D faces. Also, it is a hard and time-consuming task to create dataset using infrared laser beams or 3D scanning. As a result, Bosphorus database\cite{Bosphorus} contains 4,652 scans of 105 individuals and CASIA-3D FaceV1 dataset \cite{BID} contains 4,624 scans of 123 individuals. Researches like \cite{Millions-3D-dataset}, \cite{3DMM} provide ways to generate 3D dataset form 2D dataset.

\subsection{Paper summary} %- overall summary -- delowar
\hl{We have presented various types of deep learning architecture of face recognition system}. In this paper,\hl{ we have discussed different types of models, datasets, loss functions, and lots of occlusion handling techniques for FR task.} A taxonomy of the deep learning methods, loss functions and activation functions used for Face Recognition is shown in Figure \ref{fig:face_recog_taxonomy}. Figure \ref{fig:year-based_distribution} shows the year-based distribution of the discussed papers and
\hl{the most recent DL-based tasks for face recognition system have been discussed.} Figure \ref{fig:deep_learning_architecture} shows the discussed DL methods of this paper. \hl{In the DL-based FR system, we have found that CNN plays an important role.} Many types of FR tasks have been proposed based on the CNN model. Some of them are ResNet50, LightCNN-v9, SqueezeNetResNet-50, VGG16 and so on \cite{guo2019survey}.
Other Deep Learning-based algorithms such as Autoencoder (AE), Generative Adversarial Networks, Deep Belief Networks, Hybrid Networks, and Deep Reinforcement Learning \cite{liu2020deep} have been briefly discussed in this paper. \hl{A Table has been constructed that merged the DL-based FR tasks with datasets, architectures, and accuracy.} 
\par 
Datasets are an essential factor in a machine learning system. \hl{DL algorithms can not do their job according to the user requirements  without sufficient features in the datasets.} LFW, YTF, YTC, IJB-A, IJB-B, IJB-C, CASIA-WebFace, MS-Celeb-1M, IMDb, VggFace2 and Celebrity-1000 datasets \cite{guo2019survey} are vastly used to train the DL-based FR system and test the performance of the model.
We have also categorized some activation functions that are generally used in FR tasks. Most of these are Sigmoid, Tanh, Softmax, ReLu, Softplus, Leaky RelU, Parametric ReLU, ELU, Swish and Maxout \cite{apicella2021survey}. Moreover, we have discussed Hierarchical Softmax Loss, Contrastive Loss, Triplet Loss, N-pair Loss, Marginal Loss, Ring Loss, COCO Loss and Softmax Loss \cite{guo2019survey}.
Still image-based datasets and video-based datasets \hl{are} also discussed. We have also took the most popular LFW dataset for comparing the accuracy of various models.
In conclusion, recent Deep Learning-based face recognition methods have been thoroughly discussed in our paper.

\subsection{contributions}
% ---- delowar
The highlighted points of this paper are noted below:
\begin{itemize}
\item \hl{Recent works on DL-based FR tasks have been discussed in our paper.}
\item Briefly introduce the new DL-based FR Models. e.g., Deep Belief Network, Deep Reinforcement Learning.
\item Detail discussion of the loss functions and activation functions.
\end{itemize}

\subsection{organization }
%-- delowar -- only section not subsection.
We have organized the rest of this paper in the following manner.
\begin{itemize}

\item \textbf{\ref{sec:Deep Learning Method} Deep Learning Methods}: Face recognition models based on Deep Learning. 
\item \textbf{\ref{sec:AllOverTable111} Comparison of Different Deep Networks}: Comparison of Different Deep Networks by accuracy on a dataset. 
\item \textbf{\ref{sec:Loss function and Activation function} Loss functions and Activation functions}: Different types of Loss functions and Activation functions are discussed in this section.
\item \textbf{\ref{sec:Challenges in Face Recognition Using Deep Learning} Challenges in Face Recognition Using Deep Learning}: Occlusion and other various types of challenges has been discussed.
\item \textbf{\ref{sec:Face Datasets} Face Datasets}: Most used still images and video datasets for Face Recognition tasks have been discussed in this section.
\item \textbf{\ref{sec:Future Trends} Future Trends}:\hl{Various types of application of DL-based FR tasks with the direction of Future Trends have been discussed here.}
\item \textbf{\ref{sec:Conclusion} Conclusion}: Overall  summary of our work.
\end{itemize}

\section{Deep Learning Methods}
\label{sec:Deep Learning Method}
%1. basic introduction
%2.Visualization - later
%3. Table format - Algorithm, summary, dataset used, performance
\label{sec:units}
\subsection{Convolutional Neural Network} %- Delowar, Awal
\begin{figure}[ht]
     \centering
    \includegraphics[width=\linewidth]{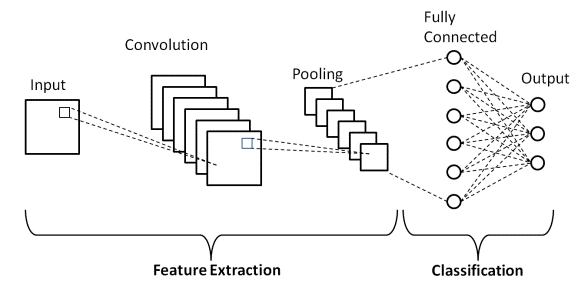}
    \caption{Basic CNN Architecture \cite{CNN_img_phung2019high}.}
    \label{fig:basic_cnn_arc_img}
\end{figure}

%--delowar \\
Convolutional Neural Network \cite{CNN_Lecun} is the most popular Deep learning algorithm for image recognition, image classification, pattern recognition, and other feature extraction operation from an image\cite{CNN_CosKun}. There are many types of CNN algorithm. But basically, two types are presented here to explain the CNN algorithm. One is feature extractor and the other is the classifier. The name of CNN comes from a mathematical linear operation between two matrices known as convolution. In CNN, one matrix is the image and the other is the kernel (operator). Actually, an image is a simple single channel (gray-scale image) or three-channel (colour image) matrix, each entry in this matrix is one pixel of the image. The dimension of the image matrix is (HxWxD). Here, H=height, W=width and D=RGB channel for RGB colour image. The Grayscale image channel contains one and the colour image channel contains three RGB colour channel. The kernel (operator) is also a matrix that has dimension of (MxNxD). Here M and N are arbitrary but most popular kernels such as Edge detectors or other operators use 3x3 size kernel. Here, D means the depth or dimension of the kernel. The dimension of the kernel is similar to the image colour channel. Figure \ref{fig:basic_cnn_arc_img} describes the architecture of CNN. In Face Recognition systems, CNN shows an excellent performance and many models have been built from CNN architecture. Table \ref{tab:CNN_summary} shows overview of CNN-based face recognition systems.

The basic CNN architecture contains four layers: the convolutional layer, pooling layer, non-linear, and fully-connected layer. The first two layers are parameterized, and the other two are non-parameterized \cite{CNN_basic}.
Parameterized layers are convolutional layers and fully-connected layers. However, non-parameterized layers are nonlinear layers and pooling layers. However, this architecture may change according to the problem requirements.
After modifying the CNN architecture, the researchers build many FR architectures, e.g. VGGNet, GoogLeNet, and ResNet, etc.\cite{guo2019survey}
Most deep face recognition systems work in a supervised fashion\cite{single_deep_CNN}.
\par 
Resnet50 is a CNN-based architecture that contains 50 layers and used for the supervised learning. A new dataset has been created manually named VGGFace2\cite{cao2018vggface2}, which was used to train ResNet50 and SqueezeNet-ResNet-50\cite{guo2019survey} models.
Wen and Li.\cite{Latent_Identity} proposed a novel approach for age-invariant face recognition. The author claimed it as the first deep CNN-based age-invariant work. They used the largest dataset of age-invariant for training and tested and discussed the accuracy on various datasets. 
\par
%overview of all summary of CNN model
% \FloatBarrier
\begin{table*}[ht]
 \begin{center}
 \caption{Overview of CNN based Deep learning Models.}
    \label{tab:CNN_summary}
\begin{tabular}{p{1in}p{1in}p{1in}p{3in}}
\hline
 \textbf{Algorithm}  & \textbf{Dataset} &  \textbf{Accuracy (\%)} &\textbf{Description}\\
 \hline\hline
ResNet50 \cite{cao2018vggface2}&IJB
& 0.995±0.001 & A large-scale dataset named as VGGFace2 has been created and the authors compared its performance with  CelebFaces+, LFW , and MegaFace datasets using the ResNet50 model.
\\
SqueezeNet-ResNet-50 \cite{cao2018vggface2}& IJB & 0.996±0.001 &This model shows excellent performance on pose and age variation.
\\
ResNet-64\cite{ring_loss}& IJB-A  &93.22 & An elegant normalization approach for deep neural network called Ring loss. 
\\
ResNet-27\cite{deng2017marginal} & LFW & 99.48 & The marginal loss simultaneously minimises the intra-class variances as well as maximises the inter-class distances by focusing on the marginal samples.
\\
LightCNN-v29\cite{chen2020identity} & LFW &98.98   & A low-resolution face recognition (LRFR) model which can perform nicely at the low resolution.
\\
 Deep CNN \cite{liu2015targeting} &LFW & 99.77 & Deep Embedding Ensemble Model with 9 Convolution layers and a Softmax layer have been used to increase the performance.
\\
MTCNN \cite{liu2017sphereface} &YTF&95 & This model can be used in hyperspherical spaces when euclidean loss can be implemented into only euclidean spaces.
\\
ReST\cite{wu2017recursive} & LFW & 93.4  & An end-to-end face identification method inspired by a spatial transformer called ReST.
\\
VGG16 \cite{peng2017reconstruction}& 300WLP & 98  & Feature learning-based face recognition model that used data synthesizing strategy to improve the accuracy on the diversity of pose.
\\
NAN \cite{yang2017neural}&YouTube Face & 95.72  & Two attention blocks have been added with the CNN model.
\\
DDRL\cite{DDRL_FV_FI} & YTF & 94.2  & Contains two parts a DCNN based encoding network and a distance metric module.
\\
PDA\cite{wang2020hierarchical}& VGGFace2-FP & 95.32  & Introduced multiple attention-based local branches at different scales to emphasize different discriminated facial regions.
\\
  \hline 
\end{tabular}
\end{center}
\end{table*}
% \FloatBarrier
Hu et al.\cite{hu2017attribute} presented a CNN-based process, which improved face recognition performance. They added an extra layer with CNN, which was equivalent to Gated Two stream Neural Network (GTNN).
Here, the authors proposed a robust nonlinear tensor-based fusion framework for face recognition, which can optimize Face Recognition Feature (FRF) and Face Attribute Feature (FAF) using low-rank tensor optimization and a GTNN. First, they systematically investigated and verified the various face recognition scenario, such as pose and illumination. Then, the authors applied a low-rank Tucker-decomposition of a tensor-based fusion framework, equivalent to GTNN, optimized by a neural network. The authors got 99.65\% accuracy on the LFW dataset and 99.94\% on CASIA NIR-VIS2.0 (Cross-modality environment). However, 100\% accuracy was achieved using ±45º pose angle.

Face recognition accuracy has been improved using a pre-trained model of VGG-Face net and Lightened CNN\cite{mehdipour2016comprehensive}. A comprehensive analysis has been done based on some occlusion conditions. These conditions are upper and lower face occlusion, varying head pose angles, misalignment due to erroneous facial feature localization, and changing illumination of different strengths.
Five popular datasets are used in this experiment. These datasets are AR face database, CMU PIE, Extended Yale dataset, Color FERET database and FRGC database.
The authors\cite{mehdipour2016comprehensive} claimed that the FaceNet model achieved 95.12\% accuracy on the YTF dataset and 99.63\% accuracy on the LFW dataset. Again, applying the DeepFace Network increases the LFW datasets' accuracy, which is 97.35\% and the accuracy for the YTF dataset is 91.4\%. DeepID network was trained on the Celebrity Faces dataset (CelebFaces) and tested on the LFW dataset, and achieved an accuracy of 97.45\%. After modifying some architecture of the VGG-Face net and Lightened CNN model, the performance was also evaluated. Some factors such as illumination, occlusion, misalignment, and head pose had reduced the face recognition accuracy of the Deep learning model. In these cases, VGG-Face has achieved better performance than Lightened CNN. Five popular datasets, the AR face database, CMU PIE, Extended Yale dataset, Color FERET database, and the FRGC database, had been used in this experiment.
\par
A new face identification-verification technique using a Deep Convolution Network was proposed, known as Deep IDentification-verification features (DeepID2)\cite{parkhi2015deep}, by increasing inter-personal variations and reducing intra-personal variations of images. LDA, Bayesian face, and unified subspace models have limitations. These models are developed to handle inter and intra-personal variations. However, when the variations are more complex, these models show limited performance \hl{and achieved} 99.15 ± 0.13\%  accuracy gain on the LFW dataset.
\par
Chen et al.\cite{chen2020identity} proposed a low-resolution face recognition (LRFR) model. LRFR can perform creditably at the low resolution of the face smaller than 32*32 pixels. The authors gave priority to both angle discrepancy and magnitude discrepancy or magnitude gap between high resolution (HR) and corresponding low resolution (LR) face pairs. The purpose of the article was to recover the identity-aware information for LRFR. LR faces increase the angle and magnitude gap of the features. The authors claimed that all super resolution-based methods reduce the angled gap and magnitude gap among the features. That is why the super-resolution network achieved 98.46\% accuracy on LightCNN-v9 and 98.98\% on LightCNN-v29, which \hl{outperformed} other renowned methods.

%-----Awal

%%%-------Auto Encoder summay-------
\begin{table*}[ht]
 \begin{center}
 \caption{Overview of Auto-Encoder based Deep learning Models.}
    \label{tab:AutoEncoder_summary}
\begin{tabular}{p{1in}p{1in}p{1in}p{3in}}
\hline
 \textbf{Algorithm}  & \textbf{Dataset} &  \textbf{Accuracy (\%)} &\textbf{Description}\\
 \hline\hline
 U-net \cite{kantarci2019thermal} & EURECOM & 88.33 & This mode deals with Illusion problem in thermal to visible cross-domain face matching.
\\
DC-SSDA \cite{cheng2015robust}& CMU-PIE & 85 & 30\% mouth occlusion and sunglass occlusion has been removed successfully from CMU-PIE face dataset.
\\
$D^2$AE\cite{8578320} & CelebA & 87.82 & Dispelling Autoencoder ($D^2$AE), a new framework that does not require previous knowledge to restore the occlusion part of the images.
\\
CAN\cite{xu2017age}&CACD-VS & 92.3 & Build the auto-encoder network that improve the face recognition performances by handling the Age variant problems.
\\
  \hline
\end{tabular}
\end{center}
\end{table*}

To perform face verification and face re-identification task, Yu et al.\cite{DDRL_FV_FI} proposed a Deep Discriminative Representation Learning (DDRL) network. It has two parts a DCNN based encoding network and a distance metric module. Here, they used $l2$ distance to verify the two images were the same or not. On the other hand, they \hl{used} softmax for face identification. Pointing out some problems with two folds face recognition, Wu et al.\cite{wu2017recursive} proposed an end-to-end face identification method. Gathering inspiration from the spatial transformer, they proposed a module called Recursive Spatial Transformer (ReST). Their model has three parts: convolution layers, localization network, and spatial transformation layer. A DCNN (modified AlexNet) network with a softmax layer follows ReST and identifies the faces from the images. They also mentioned three different types of HiReST depending on the number of hierarchies (0-2).

On the other hand, Wang et al.\cite{wang2020hierarchical} used an advanced CNN network for face recognition. They used a pyramid diverse attention to introduce multiple attention-based local branches at different scales to emphasize different discriminated facial regions at various scales automatically and adaptively. They presented a hierarchical bilinear pooling to combine features from different hierarchical layers. Lai et al.\cite{PoreNet} used the idea to detect pores from face images for face recognition called PoreNet. At first, They extracted the pore features from high regulation face image using a scale-normalized Laplacian of Gaussian (LoG) blob detector. Then they matched those features with other images to classify them. They used Grid-based Motion Statistics (GMS)\cite{gms} to reject outline. PoreNet is modified version of HardNet\cite{HardNet}. A novel approach was proposed by Wen et al. \cite{Latent_Identity} for Age Invariant Face Recognition (AIFR). This approach is a robust age-invariant deep face recognition framework. It is the first deep CNN-based age-invariant work, as they claimed. The coupled learning of latent factor-guided CNN (LIA-CNN) is beneficial to AIFR. It minimizes the classification error and maximizes the likelihood probability that latent factors generate from the training samples.

Yang et al.\cite{yang2017neural} presented a method for face identification and verification with a variable number of inputs face from image or video called Neural Aggregation Network (NAN). They extracted features from the image using CNN (GoogLeNet) network. Those features were passed through two attention blocks and assigned linear weights for them. For face verification, they used Siamese neural aggregation network and minimized average contrastive loss. Moreover, for face identification, they used a fully-connected layer followed by a softmax and minimize average classification loss.
On the other hand, Kim et al.\cite{face-body-video} mainly developed a method for face recognition from the video. This method also considers the upper body along with the face. In that paper, face detection was done by the same method as mentioned in\cite{cascade_FD} with some improvement in the network. Then they associated the body pose detected by OpenPose\cite{openpose} with face information. Finally, these two data are used for face recognition with ResFace101\cite{resnet}. 
Besides mentioning a method for augmenting the 3D face dataset, Gilani et al.\cite{Millions-3D-dataset} proposed a method to recognize them called FR3DNet. Their method maintains the same CNN-based architecture as \cite{parkhi2015deep} with some changes in convolution layers.

\subsection{Auto Encoder} %- Delowar
\begin{figure}[ht]
     \centering
    \includegraphics[width=\linewidth]{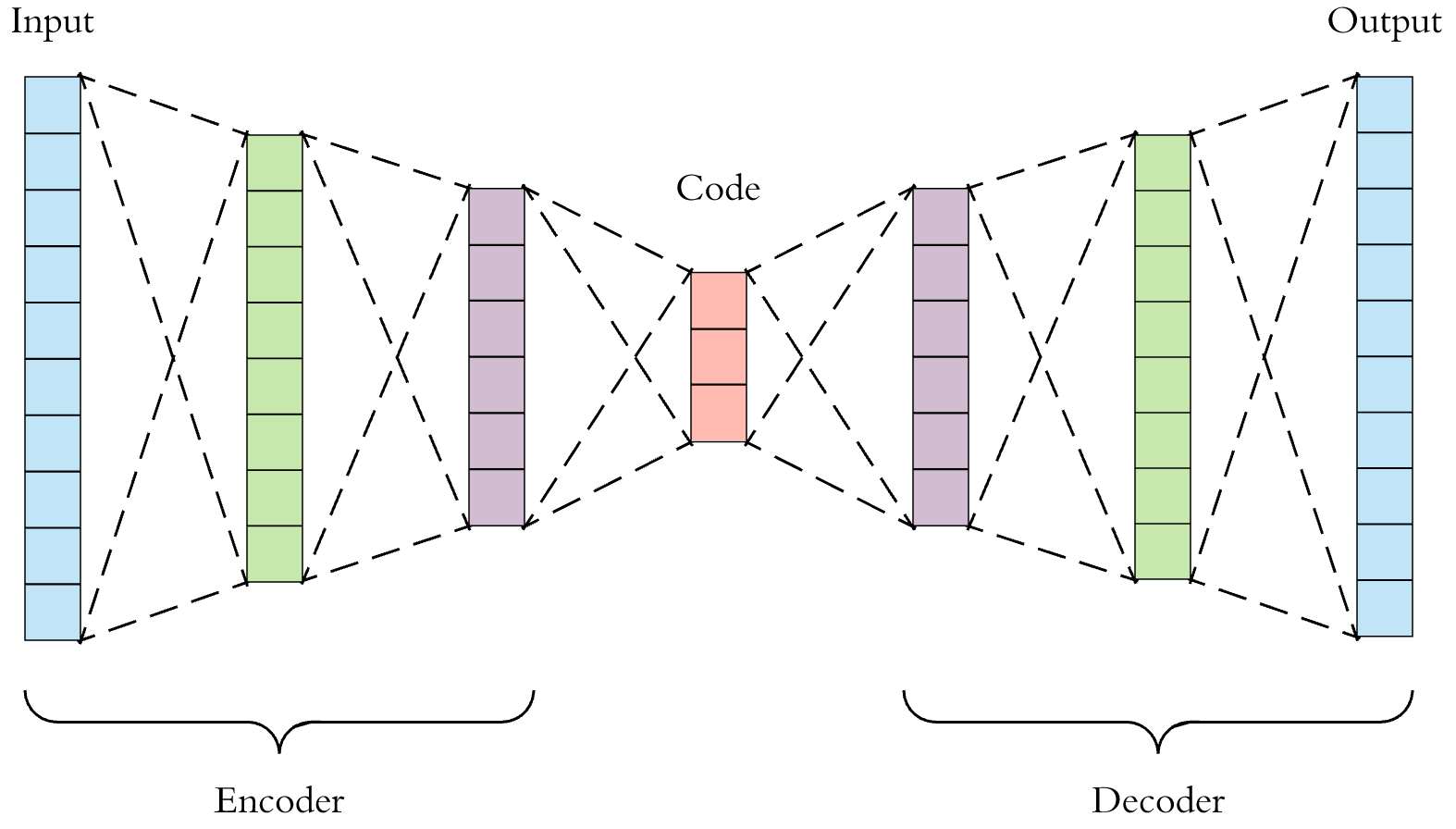}
    \caption{Basic Auto Encoder Architecture \cite{Comprehensive_Introduction_to_Autoencoders}.}
    \label{fig:AutoEncoder_architecture}
\end{figure}

Reconstructing an image from a noisy image is one of the great challenges for face recognition systems. Noisy images decrease the performance of a recognition system. Auto-Encoder is an excellent way to reconstruct a image. Table \ref{tab:AutoEncoder_summary} shows the summary of auto-encoder techniques and performance.
Auto-encoder is an unsupervised feature learning-based deep neural network which encodes and decodes the data efficiently\cite{ae_survey_deep_Alom}. 
It can automatically learn robust features from the large size of unlabeled data\cite{ae_Yuan} and for this reason, the researchers use the autoencoder to encode the input into dimension reduction and represent it with significance.  This technique has two stages: one is encoding, and the other is decoding. The entire architecture contained one or more hidden layers with an input and an output layer. In the encoding stage, the input compresses into a lower-dimensional feature with a meaningful representation. This process is continued until the required dimension is achieved. The next stage is the decoding phase. In this phase, the process is reversed to generate essential features from the encoded stage. The back-propagation is applied at the time of training models. The setting is set in the layer-by-layer decoding stage according to the input target size; thus, the error can be minimized. It decodes again, reconstructing the output similar to the original input. There are many variations in autoencoder techniques, for example, the denoising autoencoder\cite{wang} technique was proposed to improve the image representation ability of the autoencoder. A basic architecture is shown in Figure \ref{fig:AutoEncoder_architecture}.

Though autoencoder combines the generative and learning properties to learn in an unsupervised manner, sometimes it can learn disentangled representations. Adversarial Latent Autoencoder (ALAE) \cite{ALAE_Pidhorskyi} was proposed to handle this type of limitation. ALAE architecture improves the training procedure of GAN. Manifold-value data comes from the medical image. Higher-dimensional data arise when Magnetic Resonance Imaging (MRI) on brain connectomes in cognitive neuroscience. This higher dimensional space cannot reduce using PCA and for this reason, uncertainty arises when analyzing manifold-value. To overcome this situation Miolane et al.\cite{Miolane_2020_CVPR} proposed a Riemannian variational autoencoder but in low light condition, face recognition models can not perform properly. To solve the problem, Face matching in cross-domain thermal to visible techniques \hl{have} been proposed. A deep autoencoder\cite{kantarci2019thermal} based method learns from mapping between thermal and visible face images. Extensive works have been done using the Deep autoencoders and facial expression that can recognize images by reducing the dimensions\cite{DBLP}.
\par 
The age-invariant problem in face recognition systems was solved by a couple of autoencoder-based face recognition techniques (CAN)\cite{xu2017age}. CAN is constructed from two autoencoders and performed well for aging and de-aging on the complex nonlinear process using two shallow neural networks.

%%------GAN--------------
%--------------------------
\begin{table*}[ht]
 \begin{center}
 \caption{Overview of GAN based Deep learning Models.}
    \label{tab:GAN_summary}
\begin{tabular}{p{1.2in}p{1in}p{1in}p{3in}}
\hline
 \textbf{Algorithm}  & \textbf{Dataset} &  \textbf{Accuracy (\%)} &\textbf{Description}\\
 \hline \hline
 R3AN\cite{r3an} & MegaFace & 98.46 & For cross model FR problems and experimented on public datasets. \\
 
 ACSFC\cite{he2019adversarial} & CASIA NIR-VIS 2.0 & 99.8±0.09 & Improved the performance of NIR-VIS heterogeneous samples.\\
 
 AFRN-GAN\cite{du2019age} & MULTIPIE & 95.7 & Combination of transfer learning and adversarial learning, proposed for cross-age FR. \\
 
 Age-cGAN\cite{antipov2017face} & IMDB-Wiki cleaned & 82.9 & Improved the face aging system but can't improve the face verification problem.\\
 
 Age-cGAN+LMA\cite{antipov2017boosting} & LFW & 88.7 & Solve the drawbacks of Age-cGAN, and perform better in face verification.\\
 
 DA-GAN\cite{zhao2017dual} & IJB-A & 99.1 ± 00.3 & A face synthesis based model for extreme poses. Projects 3D face image to 2D using the learned knowledge from GAN.\\
 
 TR-GAN\cite{kezebou2020tr} & Tufts Face & 88.65 & A cross model over thermal to RGB. GAN is used mainly for training the loss function.\\
 
 DR-GAN\cite{tran2017disentangled} & CFP & 93.41 ± 1.17 & Solved the large pose variation. Performed well on both single and multi-image. \\
 
 CpGAN\cite{iranmanesh2020coupled} & CASIA NIR-VIS 2.0 & 96.63 & For heterogeneous FR and Used perceptual loss in the model.\\
 
 FI-GAN\cite{FIgan} & LFW & 99.6 & Improved the performance on large face pose based images. \\
 
 PA-GAN\cite{liu2020pa} & IJB-A & 99.0 ± 0.2 & Increase the accuracy on raw surveillance of FR. Discriminativeness of a face is enhanced.\\
 \hline 
\end{tabular}
\end{center}
\end{table*}
%---- gan figure
\begin{figure}[htbp]
    \centering
    \includegraphics[width=86mm,scale=1.0]{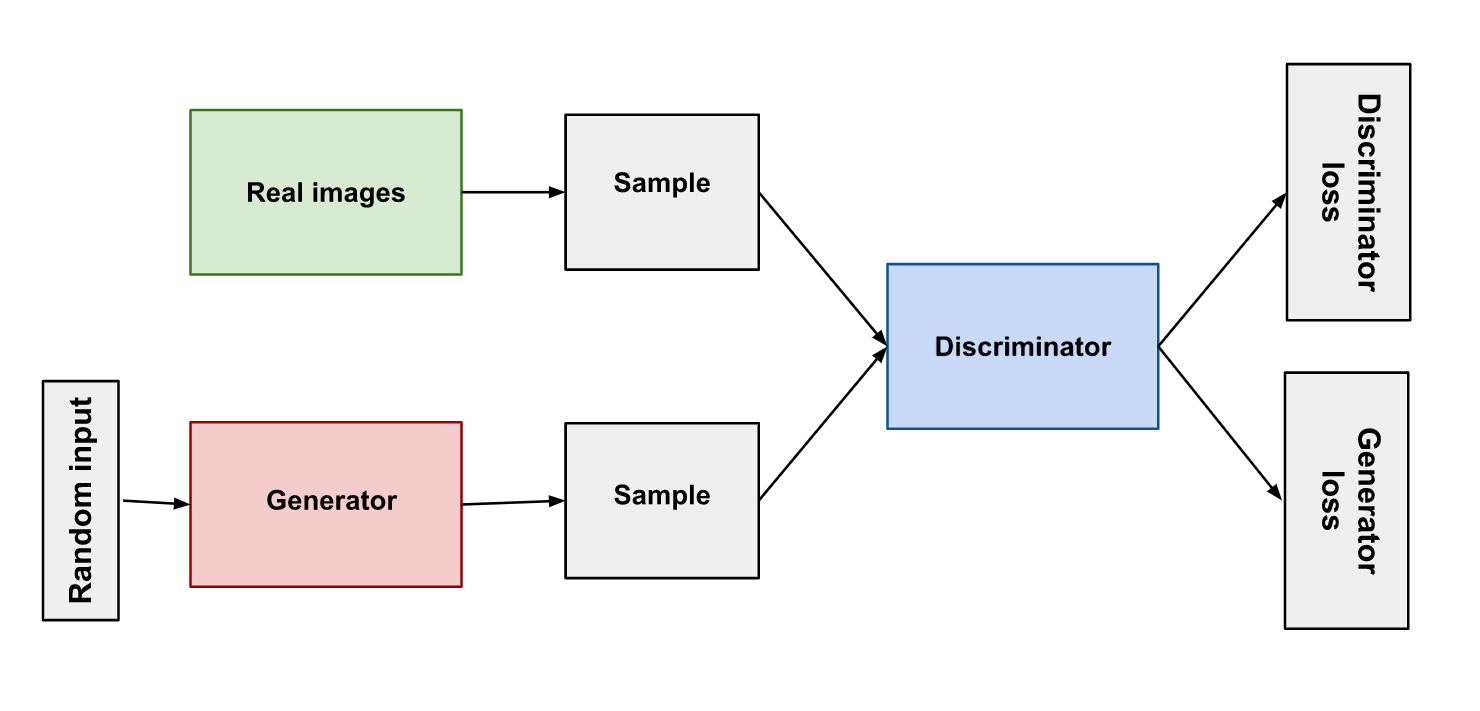}
    \caption{Basic GAN Diagram \cite{google}.}
    \label{fig:GAN_diagram}
\end{figure}
%---------
\subsection{Generative Adversarial Network}%- Fuad\\
Generative Adversarial Networks are another kind of unsupervised deep learning method. It automatically discovers and learns the regularities or patterns from the input data. Figure \ref{fig:GAN_diagram} shows the block diagram of a GAN. The GAN model includes two sub-models: a generator model for generating new features and a discriminator model for classifying whether generated features are actual, taken from the domain, or fake, generated by the generator model. GANs are based on a game-theoretical schema where the generator network has to contend against an adversary. The generator part generates features and examples directly from its adversary, and the discriminator part tries to differentiate among the samples taken from the training data and the samples taken from the generator\cite{goodfellow2016deep}. Table \ref{tab:GAN_summary} describes the overview of GAN methods. 

GANs are utilized in solving general face recognition problems as cross-age face recognition, face synthesis, pose-invariant face recognition, video-based face recognition, makeup-invariant face recognition, and so on. For example, R3AN architecture\cite{r3an} was proposed for cross model FR problem. It divides the method into three paths: reconstruction, representation, and regression for training. Moreover, using a mapping function, it maximizes the conditional probability. TR-GAN\cite{kezebou2020tr} was proposed as a cross model over thermal to RGB. Here, GAN is used for loss training, and the generator part synthesizes images with fine details. For improving the performance of NIR-VIS heterogeneous samples, ACSFC\cite{he2019adversarial} was proposed. It prototypes a high-resolution heterogeneous\hl{\mbox{\cite{ouyang2016survey}}} face synthesis with two components: a texture inpainting component and a pose correction component. A novel 3D pose correction loss, two adversarial losses, and a pixel loss are used for generating results. DA-GAN\cite{zhao2017dual} was proposed to do face synthesis under extreme poses. It merges the knowledge from adversarial training and domain from perception losses and projects 3D face image into 2D face image space.

A novel framework AFRN-GAN\cite{du2019age} was proposed for cross-age face recognition. It combines transfer learning (TL) and adversarial learning (AL). The discriminator part is trained to discriminate the age information, and the generator part extracts features using TL and suppresses age information using AL. Age-cGAN\cite{antipov2017face} was proposed to improve the aging system. However, it fails to improve much in the face verification sector. For overcoming this, Age-cGAN+LMA\cite{antipov2017boosting} was proposed. This combination improves the drawbacks of Age-cGAN. An encoder-decoder structure-based Disentangled Representation learning-Generative Adversarial Network, DR-GAN\cite{tran2017disentangled}, was proposed to solve the large pose variation. Here, generator takes a face image, a pose-code $c$, and a random noise vector $z$ as the inputs to generate a face of the same identity with the target pose that can fool the discriminator. It works on both single image and multi-image. CpGAN\cite{iranmanesh2020coupled} was proposed for heterogeneous face recognition. This model has two sub-networks; each has a separate GAN. The first sub-network is for the visible spectrum, and the other one is for the non-visible spectrum. The authors used a dense encoder-decoder structure with multiple loss functions to keep the features from each sub-network close to each other. They also used perceptual loss function in the coupling loss function.

\begin{figure}[ht]
     \centering
    \includegraphics[width=\linewidth]{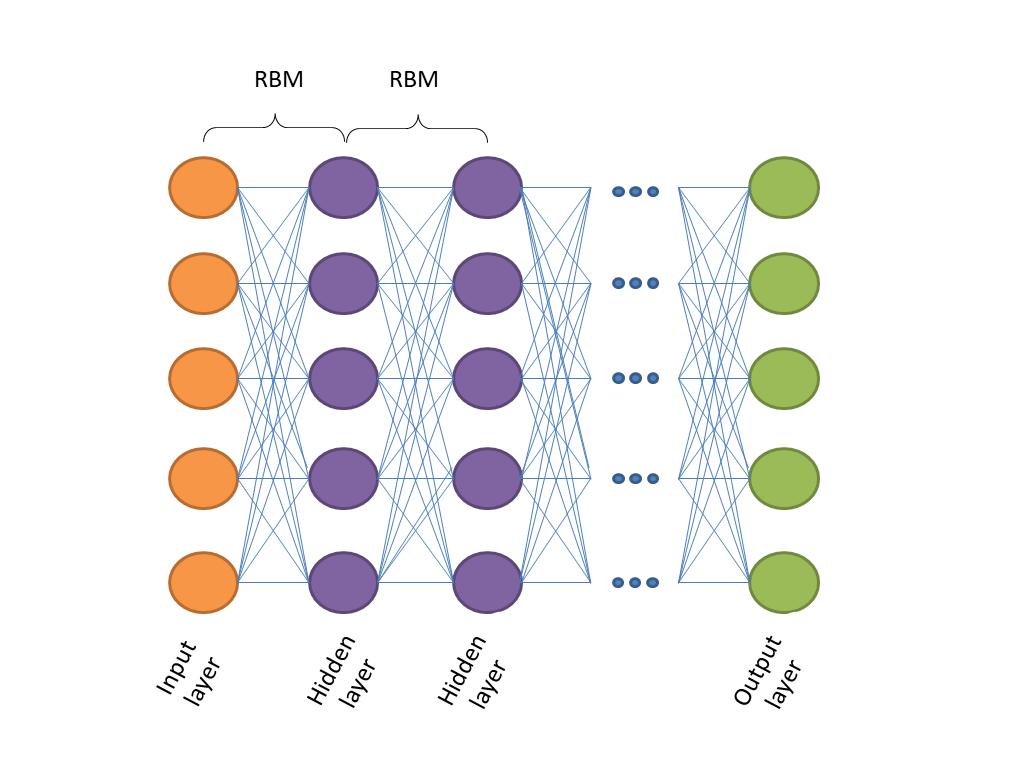}
    \caption{Deep Belief Network Architecture.}
    \label{fig:basic_auto_encoder_img}
\end{figure}

\subsection{Deep Belief Network}% -- Awal

Traditional DNN-based networks have some problems such as stuck at local optima, slow learning and require a lot of training datasets. A type of DNN was proposed to cope with those problems, a composite of multiple hidden units called Deep Belief Network (DBN) \cite{DBN}. In DBN, hidden units of different layers are internally connected, but the units of the same layers are not connected. It can be treated as a sequence of restricted Boltzmann machines (RBMs) or autoencoder where each hidden sub-layer works as a visible layer for the next hidden sub-layer. It generally ends with a softmax layer for classification. Figure \ref{fig:basic_auto_encoder_img} describes the Deep Belief Architecture. Table \ref{tab:DBF_summary} shows a brief overview of the deep belief network works on face recognition.

Taking advantage of convolutional restricted Boltzmann machines (CRBM), Huang et al.\cite{local_CRBM} developed a novel method to learn face recognition features. It is a local CRBM and applied to high-resolution images. They cropped the images into three different sizes and used them as inputs. Then, they divided the images into some overlapping regions and assigned a different set of weights for a different region. After that, they applied Information-Theoretic Metric Learning (ITML)\cite{ITML} to produce a Mahalanobis matrix\cite{mahalanobis_metric}. Finally, a linear SVM\cite{svm} was applied to perform face verification.
%------DBF table
\begin{table*}[ht]
 \begin{center}
 \caption{Overview of Deep Belief Network based Models.}
    \label{tab:DBF_summary}
\begin{tabular}{p{1.2in}p{1in}p{1in}p{3in}}
\hline
 \textbf{Algorithm}  & \textbf{Dataset} &  \textbf{Accuracy (\%)} &\textbf{Description}\\
 \hline\hline
 Fan et al.\cite{DBN_dropout} & ORL & 93.5 & Deep belief network with 2 hidden layer\hl{s} and 500 nodes each and 50\% dropout.\\
 
 Huang et al.\cite{local_CRBM} & LFW  & 87.77 ± 0.62 & Novel local convolutional Restricted Boltzmann Machines, Information-Theoretic Metric Learning, Cholesky decomposition, SVM.\\
 
 Annamalai and & ORL & 98.92 & Feature extraction using BRISK and LTP and \\
 Prakash\cite{firefly_DBM} & YALE & 97.92 &  optimization with enhanced fire-fly.\\
 
 Bouchra et al.\cite{DBN_compare} & BOSS & 98.86 & 4 layers of neural network for DBN.\\
 & MIT & 98.04 & \\
 \hline 
\end{tabular}
\end{center}
\end{table*}
%------
Fan et al.\cite{DBN_dropout} used DBN on a small Olivetti Research Laboratory (ORL)\cite{orl} dataset with two hidden layers with 500 units per layer for face recognition. As the dataset is small, the model might overfit. So, they randomly added 50\% dropout to reduce that. As a result, their model achieved high accuracy, as they claimed. However, they did not mention how the dropout would react on a larger dataset.

On the other hand, Annamalai\cite{firefly_DBM} divided the face recognition task into five sub-steps: image collection, image de-noising, feature extraction, optimization, and classification. The collected images are from ORL\cite{orl}, YALE\cite{yale}, and Face Semantic Segmentation (FASSEG)\cite{FASSEG} databases. They used Histogram equalization for image de-noising, Local Ternary Pattern (LTP)\cite{ltp} and Binary Robust Invariant Scalable Keypoints (BRISK)\cite{brisk} for feature extraction and enhanced fire-fly\cite{firefly_optimization} for optimization. Finally, a DBN network is used to classify the facial images.
Bouchra et al.\cite{DBN_compare} \hl{made} a comparison between three models on BOSS\cite{DBN_compare} and MIT\cite{MIT_D} dataset. Those models are DBN with four layers, Stacked Auto-Encoder (SAE) with three layers, and Back Propagation Neural Networks (BPNN) with three layers. DBN outperformed the other two models on both datasets. DBM is sometimes used for liveness detection as an extended part of face recognition\cite{DeBNet}.

%%%%%------------hybrid
\begin{table*}[!htbp]
 \begin{center}
 \caption{Overview of Hybrid Models based Face Recognition.}
    \label{tab:hybrid_summary}
\begin{tabular}{p{1in}p{1in}p{1in}p{3in}}
\hline
 \textbf{Algorithm}  & \textbf{Dataset} &  \textbf{Accuracy (\%)} &\textbf{Description}\\
 \hline\hline
 CNN+RBM \cite{sun2013hybrid} & LFW & 91.75
 & Deep ConvNets find the relation of a feature and Restricted Boltzmann Machine calculates the inference of feature extraction data.
\\
 DBMs + AEs \cite{DBMs_AEs} & WhoIsIt &28.53 &Significantly increases the face recognition performance on weight and age variations.
\\
 MDLFace \cite{DBM_SOD} &  PaSC & 93.4 &Face recognition from Video datasets.
\\
 CAN \cite{CNN_GAN_CAN} &  FRGC Ver2.0 & 97.92 & A cross modal based Deep learning method that can perform heterogeneous matching between depth and color of images.
\\
 DeepID3 \cite{sun2015deepid3} & LFW & 99.52 & VGG-net and GoogLenet Based hybrid model for Face Recognition.
 \\
 CNN and deep metric learning \cite{liu2015targeting} &LFW&99.77 & This system performs on discriminative low dimensional features that improve the performance of face recognition.
 \\
 
  \hline 
\end{tabular}
\end{center}
\end{table*}

\subsection{Hybrid Network}% - Delowar \\
A hybrid model is a combination of two or more generic machine learning models to improve the overall performance of the model. In this technique, one algorithm augments another algorithm to solve problems precisely. Most of the time, a single machine learning algorithm is designed for a particular task. However, when two or more algorithms are combined, the performance of the hybrid model significantly increases. Some hybrid models are CNN+GAN, CNN+AE, GAN+RL, etc. Table \ref{tab:hybrid_summary} shows the summary of a Deep learning-based hybrid \hl{models} for the face recognition system.

Sun et al.\cite{sun2013hybrid} proposed a combining ConvNet Restricted Boltzmann Machine for face verification. Generally, when the dataset comes with a high dimension and more complex feature vector, the dataset is needed to be compressed for feature extraction. Hybrid models are more robust process than a single algorithm for extracting features. CNN extracts features from two images; hence we can compare that both are similar or not. Here, the RBM method calculates the inference of image features to overcome the complexity.
Sing et al. combined DBMs and AEs for face recognition\cite{DBMs_AEs}. This model follows the regularize-based process that easily learns facial-invariant problems. This method improves accuracy significantly.
Goswami et al.\cite{DBM_SOD} proposed a hybrid model named MLDFace. It is a combination of DBM and a stack of Denoising Autoencoders for the video-based face recognition framework. Another face recognition hybrid model, Conditional Adversarial Networks \cite{CNN_GAN_CAN}, was proposed to combine DCNN and GAN for cross-modality learning.\par
Instead of using traditional handcrafted features such as LBP or HOG, Liu et al.\cite{liu2015targeting} introduced a two-stage face recognition method. It shows high-performance in the real-world face recognition system. Multi-patch deep CNN and deep metric learning methods are combined to build this model. This method can recognize faces with variant poses, occlusions, and expressions correctly. However, the number of faces and identities, data size, and the number of patches in training data are crucial for achieving the final performance. After a certain number of patches, the error rate of test data increases due to overfitting issues. When the authors combined ten models and train the data with that combined model, it showed the best result.
%----table of RL
\begin{table*}[ht]
 \begin{center}
 \caption{Overview of Deep RL based Models.}
 \label{tab:RL_summary}
 \begin{tabular}{p{1.2in}p{1in}p{1in}p{3in}}
 \hline
 \textbf{Algorithm}  & \textbf{Dataset} &  \textbf{Accuracy (\%)} &\textbf{Description}\\
 \hline \hline
  Fair Loss\cite{liu2019fair} & LFW & 99.57 & Learns margin adaptive strategy to make the additive margin more reasonable and solves the class imbalance problem. \\
 
  ADRL\cite{adrl} & YTF & 96.52 ± 0.54 & For video FR and use MDP for finding the attention of videos.\\ 
 
  RL-RBN\cite{wang2020mitigating} & RFW\cite{wang2019racial} & 95.79 & Reduce racial bias by using RL. \\
 
  AFA\cite{duong2019automatic} & AGFW-v2 & 83.67 & Face aging technique to generate a future face in old age from a young face in a video frame.\\
  Xiaofeng et al.\cite{liu2019dependency} & IJB-A & 0.976±0.01 & Focuses on set-based face verification and uses actor-critic reinforcement learning to create a dependency-aware attention control network. \\
 \hline 
 \end{tabular}
 \end{center}
\end{table*}
%---------

Yang et al.\cite{yang2017neural} proposed Neural Aggregation Network (NAN) for video database based face recognition. This hybrid network is made using GoogleNet and Siamese neural aggregation networks. The authors extracted features from the image using CNN (GoogleNet), which passed through two attention blocks. For face verification, they used Siamese neural aggregation network and minimized average contrastive loss. For face identification, they used a fully connected layer followed by a softmax and minimize average classification loss. 
A hybrid network was proposed combining VGG-net and GoogLenet, named DeepID3\cite{sun2015deepid3}, that improved face verification and identification accuracy using very DNN architecture. DeepID3 network is rebuilt from VGG-net and GoogLenet to change their exterior architecture. DeepID3 network shows excellent performance on LFW faces in verification and identification. When training on large-scale data, the efficiency will be increased. When the face labels are accurate, the accuracy of DeepID3 is 99.52\% on the LFW dataset and 96.0\% on the LFW Rank-1 dataset. DeepID3 net1 and DeepID3 net2 reduce the error rate by 0.81\% and 0.26\% compared to DeepID2+ net.  

%%%%----------Deep Reinforcement Learning-----
%%%%%%%%%%%
\subsection{Deep Reinforcement Learning}% - Fuad\\

\begin{figure}[htbp]
    \centering
    \includegraphics[width=86mm,scale=1.0]{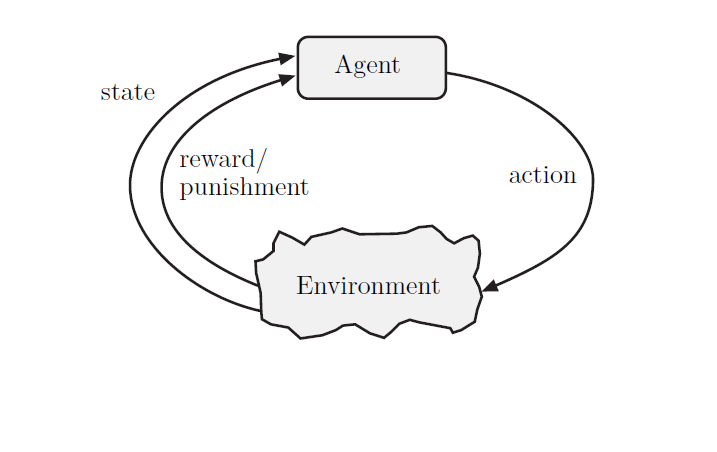}
    \caption{Basic RL Diagram \cite{tizhoosh2006reinforced}.}
    \label{fig:RL_diagram}
\end{figure}

Reinforcement Learning learns from the adjacent environment. It originated from humans' decision-making procedure\cite{littman2015reinforcement}, enabling the agent to decide the behavior from its experiences by trial and error. Figure \ref{fig:RL_diagram} describes the basic Reinforcement Learning state diagram. The combination of Deep Learning and Reinforcement Learning is used mainly in face recognition. Researchers use it in various sectors of Face Recognition. Table \ref{tab:RL_summary} shows a brief overview of RL in the face recognition method.

In RL-RBN\cite{wang2020mitigating}, the racial bias of face recognition has been reduced. The authors also proposed an optimal margin loss for this model. The authors had created two train datasets and applied the RL-based race-balance network. They also used the Markov decision process (MDP) to find the optimal margin for non-Caucasians. They worked on the self-created dataset. Chi et al.\cite{duong2019automatic} used Deep RL for the face aging technique. It generates a future face of old age from a young face in a video frame. Here, the first step is to take feature embedding using CNN and normalize using VGG-19 with an additional extra conv3\_1, conv4\_1, conv5\_1 layers. In this model, deep RL is used for neighbor selection. It can exploit the temporal relation among two consecutive frames. Here, each 900x700 resolution video frame needs 4.5 minutes.

In Fair Loss\cite{liu2019fair}, the authors used Deep Q-Learning to train an agent to learn a margin-adaptive strategy to make the additive margin more reasonable for various classes. Moreover, it solves the class imbalance problem. In this model, first, they trained a CNN by manually changing the margin in the loss to collect a series of samples. Then, the samples were used to train an agent for the margin-adaptive strategy. Finally, they trained fair loss networks with margins changing by the action outputs from the agent. They used a two-layer fully connected network with proposed Q-function; a ReLU activation function follows each layer. 
\par

For video face recognition, an attention-aware deep reinforcement learning (ADRL)\cite{adrl} was proposed. The authors made a system of finding videos' attention as a Markov decision process and used deep RL for training the attention model. They took a pair of face videos as the input of the attention model. This framework has two parts: feature learning and attention learning. The first part is processed with a deep CNN model, a recurrent layer, and a temporal pooling layer, and the second part is a frame evaluation network, which produces the values of the frames. They introduced a flexible local bi-directional recurrent layer and a local temporal-pooling layer using long short-term memory (LSTM). They tried to adopt a human strategy using reinforcement learning to remove the worst \hl{frames} step by step. The remaining frames are the most sensitive ones.

\section{Comparison of Different Deep Networks}
\label{sec:AllOverTable111} %Allover table - Iftee

\begin{figure}[ht]
\centering
\begin{tikzpicture}[scale=0.95]
\begin{axis}[
    xlabel={Year},
    ylabel={Accuracy},
    xmin=2014, xmax=2020,
    ymin=99, ymax=100,
    symbolic x coords={2014,2015,2016,2017,2018,2019,2020},
    xtick=data,
    ytick={99.10,99.20,99.30,99.40,99.50,99.60,99.70,99.80,99.90,100.00},
    legend pos=north west,
    ymajorgrids=true,
    grid style=dashed,
]
\addplot[
    color=blue,
    mark=square,
    ]
    coordinates {
	(2014,99.15)
	(2015,99.63)
	(2016,99.50)
	(2017,99.65)
	(2018,99.76)
	(2019,99.85)
	(2020,99.89)
    };
    \legend{Accuracy}
\end{axis}
\end{tikzpicture}
\caption{Accuracy distribution by year on LFW dataset.}
\label{fig:accuracy_curve}
\end{figure}
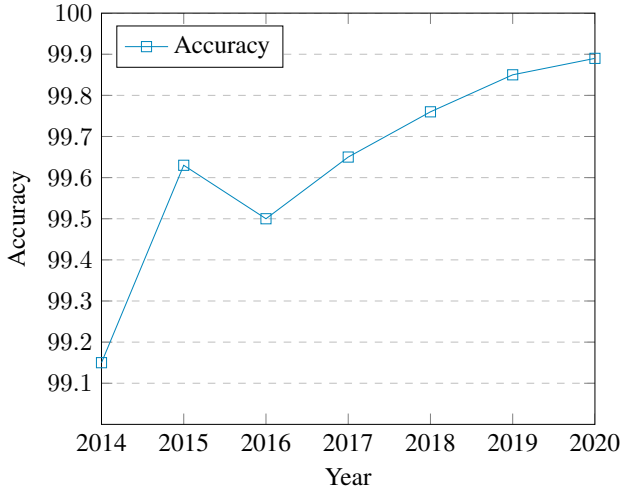

%--fuad\\
We have set LFW dataset as the benchmark dataset and compared all the proposed methods' accuracy in Table \ref{tab:AllOverTable}. \hl{In the Table}, we place the FR methods from different years. In Figure \ref{fig:accuracy_curve}, we have shown a gradually increasing graph of the performances. Though in 2016, the accuracy dropped \hl{by} 0.1\%.

\onecolumn
\begin{longtable}{p{0.9in}p{1.4in}p{0.45in}p{0.48in}p{3in}}
\caption{Verification Accuracy of the Deep Learning Methods on LFW Dataset.}\\
\hline
\label{tab:AllOverTable}
%\hline
 \textbf{Method}  & \textbf{Category or} & \textbf{Training} &  \textbf{Accuracy } &\textbf{Description}\\
 \textbf{} & \textbf{Architecture} & \textbf{images} & \textbf{(\%)} & \textbf{}\\
\hline\hline
\endfirsthead
\caption{Verification Accuracy of the Deep Learning Methods on LFW Dataset (\it{continued}).}\\
\hline
\label{tab:AllOverTable}
%\hline
 \textbf{Method}  & \textbf{Category or} & \textbf{Training} &  \textbf{Accuracy} &\textbf{Description}\\
 \textbf{} & \textbf{Architecture} & \textbf{images} & \textbf{(\%)} & \textbf{}\\
\hline\hline
\endhead
ConvNet-RBM\cite{sun2013hybrid} &Hybrid deep learning & 240k & 91.75 &CNN for feature extraction and \hl{the RBM method calculates} the inference of feature of two images.
\\

HPEN\cite{chal5} &3D Morphable Model&  & 95.80 &High -Fidelity Pose and Expression Normalization that automatically generates the face from frontal pose and expression.
\\
Joint Bayesian\cite{cao2013practical} &Transfer learning & - & 96.33 &Invariance in feature-transform, low computational cost, robustness, subspace learning.
\\
DeepFace\cite{chal19} &Modified Deep CNN & 2.6M & 97.35 &Training the model with a large-scale dataset of face images by a combination of automation and human with \hl{a} small noise label.
\\
Aug\cite{masi2016we}&Data augmentation & 2.4M & 98.06 &Increasing the \hl{amount of data} by data augmentation and training \hl{them} with deep CNN.
\\
Cascaded Algorithm\cite{wang2019cascaded}&LightCNN & 5M & 98.19 &Purifies noisy faces.
\\
Multibatch\cite{tadmor2016learning}&Metric learning & 2.6M & 98.20 &Reduced computational cost and increased precision.
\\
FI-GAN\cite{rong2020feature}&GAN, Feature-Mapping Block& - & 98.30 &Face recognition and frontalization  under large poses.
\\
Multimodal\cite{ding2015robust} &Modified CNN & - & 98.43 &Extracts multimodal information from \hl{the} holistic face, renders frontal pose, uniformly samples image using a set of CNN.
\\
LF-CNN\cite{wen2016latent} &MTCNN& 78k & 98.50 &Robust age-invariant deep face recognition framework, minimizes loss. 
\\
GaussianFace\cite{lu2015surpassing} &Gaussian Process & 40k & 98.52 &Multi-task learning approach and \hl{self-acting} adaption of complex data distribution form multiple source domains.
\\
L-Softmax\cite{liu2016large}&Modified loss function& - & 98.71 &Increased \hl{inter-class} variation, adjusted margin and avoidance of overfitting.
\\
VGGFace\cite{parkhi2015deep}&Modified CNN architecture& 2.6M & 98.95 &\hl{A large-scale dataset with a minor small noise label was created} and decreased annotation and fed them into a deep CNN.
\\
DDRL\cite{DDRL_FV_FI}&Siamese network& - & 99.07 &DDRL consists of an encoding network and the distance metric module.
\\
Baidu\cite{liu2015targeting}&9 Convolution layers and a Softmax layer& 1.3M & 99.15 &Deep CNN for multi patch feature extraction and metric learning for reducing \hl{the} dimensionality.
\\
DeepID2\cite{chal20}&Multi-scale Deep Convolution network& 202k & 98.71 &Increasing inter-personal variations and reducing intra-personal variations of image.
\\
NormFace\cite{wang2017normface}&L2 normalization& - & 99.15 &Cosine similarity optimized by modifying Softmax loss and metric learning brings in agent vector of each class.
\\
PSNet50\cite{srivastava2019psnet}&Modified CNN& 494k & 99.26 &Parametric Sigmoid-Norm (PSN) layer increases the gradient flow and \hl{performs} better .
\\
Center Loss\cite{center-loss}&Modified loss fuction& 0.7M & 99.28 &Finds a vector called center for deep features of each class and decreases the distance between center and deep features during training.
\\
SphereFace\cite{liu2017sphereface}&MTCNN, Advance Softmax loss& - & 99.42 &Used in hyperspherical spaces when \hl{the euclidean loss was implemented} into only euclidean spaces.
\\
DeepID2+\cite{chal17}&Modified version of multiscale Deep CNN& 290k & 99.47 &Dimensionality of \hl{the} hidden layers of ConvNets increased and added supervision to early convolutional layers.
\\
MarginalLoss\cite{deng2017marginal}&Modified loss fuction, ResNet& 4M & 99.48 &27 convolutional layers \hl{including}  batch normalization layers.
\\
\hline
Face++\cite{zhou2015naive}&Naive deep CNN& 5M & 99.50 &Mainly focused on the importance of large labelled training dataset.
\\
Range Loss \cite{zhang2017range} &Modified loss function& - & 99.52 &Finds the impact of imbalanced training dataset on deep learning model and handling methods.
\\
DeepID3\cite{sun2015deepid3}&Modified CNN& 290k & 99.53 &Architectural change in VGG net and GoogLeNet.
\\
Fair Loss\cite{liu2019fair}&ResNet50, 64-layer CNN  & - & 99.57 &Margin-aware reinforcement learning-based loss function.
\\
DMHSL\cite{yu2019deep}&Improved triplet loss fuction& 3.31M & 99.60 &A dynamic margin that \hl{decreases} the number of triplets.
\\
RegularFace\cite{zhao2019regularface}&Attention map created from low-rank bilinear pooling& 3.1M & 99.61 &Finds the identical and relational pair features from attention score of local appearance block features of faces.
\\
AdaptiveFace\cite{liu2019adaptiveface}&Modified  Softmax loss function& 5M & 99.62 &Adapted margin for various class variation and Hard Prototype Mining.
\\
FaceNet\cite{facenet}&CNN based ZF-Net and Inception architecture & - & 99.63 &Generates a high-quality face mapping from the images and  provides a unified embedding  through face image into a euclidean space.
\\
GTNN\cite{hu2017attribute}&Neural Tensor Fusion Networks& - & 99.65 &A robust nonlinear and  low-rank Tucker-decomposition of  tensor-based fusion framework.
\\
CosFace\cite{wang2018cosface}&Large margin cosine loss& - & 99.73 &A cosine margin term $m$ was introduced to maximize the decision margin in the angular space.
\\
MobiFace\cite{duong2019mobiface}&Modified CNN architecture& 3.8M & 99.73 &Deep neural network with lighter weight and low cost operator, low computational cost, high accuracy  on mobile devices.
\\
PRN\cite{kang2018pairwise}&ResNet-101& 2.8M & 99.76 &Discriminate classes by unique pairwise relation and patches of local appearance in the region of  landmark points into feature maps.
\\

URFace\cite{shi2020towards}&Confidence-aware identification loss& 4.8M & 99.78 & Handle different variation  such as low resolution, occlusion and head pose  by dividing the embedding into multiple sub-embeddings.
\\
UniformFace\cite{duan2019uniformface}&Modified loss function& 6.1M & 99.80 &Equidistributed representation in loss function  to maximize discriminability between two classes.
\\
HPDA\cite{wang2020hierarchical}&Advanced CNN& - & 99.80 &Learns multiscale diverse local representation automatically and adaptively and also reduces problems like pose variations or large expressions, or similar local patches.
\\
CurricularFace\cite{huang2020curricularface}&Adaptive Curriculum Learning& - & 99.80 &Connects positive and negative cosine similarity simultaneously without manual tuning and additional hyper-parameter.
\\
FSENet\cite{cheng2019face}&Face Segmentor& 8M & 99.82 &The local and global information of facial part utilized and structural correlation of them was built by face  segmentor.
\\
ArcFace\cite{deng2019arcface}&ResNet100& - & 99.83 & A modified loss function,
obtain highly discriminative features for FR and stabilize the training process.
\\
AFRN\cite{kang2019attentional}&Attention map created from low-rank bilinear pooling & 3.1M & 99.85 &Finds the identical and relational pair features from \hl{the} attention score of local appearance block features of faces.
\\
GroupFace\cite{kim2020groupface}&ResNet-100 and Group Decision Network & 10M & 99.85 &Instance-based Representation and Group-aware Representation \hl{which provide} self-distributed labels.
\\
NAS\cite{zhu2020new}& Neural network search and modified loss function & 5.8M & 99.89 &Combining reinforcement learning with neural network search.
\\
\hline 
\end{longtable}
\twocolumn
\hspace*{-3.3mm}
However, this comparison is based on the data from the renowned models. After this, we take methods tested in the YTF dataset and present their accuracy in Table no \ref{tab:AllOverTable3}.

%-- iftee\\

\begin{table*}[!ht]
 \begin{center}
 \caption{Verification Accuracy of the Deep Learning Methods on YTF Dataset.}
    \label{tab:AllOverTable3}
\begin{tabular}{p{1.5in}p{3in}p{1.5in}}
\hline
 \textbf{Method}  & \textbf{Training Dataset} &  \textbf{Accuracy (\%)} \\
\hline\hline

DeepID\cite{sun2014deep} & CelebFaces & 92.20 
\\
DeepFace\cite{chal19} & - & 92.50 
\\
DeepID2+\cite{chal17} & CelebFaces+, WDRef  & 93.20 
\\
Range Loss\cite{zhang2017range} & VGGs & 93.70 
\\
L-Softmax\cite{liu2016large} & MNIST, CIFAR10, CIFAR100 & 94.00 
\\
NormFace\cite{wang2017normface} & - & 94.72
\\
CenterFace\cite{center-loss} & CASIA-WebFace, CACD2000, Celebrity+ & 94.90 
\\
TBE-CNN\cite{ding2017trunk}& - & 94.96
\\
SphereFace\cite{liu2017sphereface} & CASIA-WebFace & 95.00 
\\
FaceNet\cite{facenet} & - & 95.12
\\
NAN\cite{yang2017neural}& - & 95.72 
\\
MarginalLoss\cite{deng2017marginal} & MS-Celeb-1M & 95.98 
\\
DeepVisage\cite{hasnat2017deepvisage}& MS-Celeb-1M & 96.24 
\\
PRN\cite{kang2018pairwise} & VGGFace2 & 96.30 
\\
RegularFace\cite{zhao2019regularface} & CASIA-WebFace, VGGFace2 & 96.70 
\\
AFRN\cite{kang2019attentional} & VGGFace2 & 97.10 
\\
VGGFace\cite{parkhi2015deep} & - & 97.30 
\\
CosFace\cite{wang2018cosface} & CASIA-WebFace & 97.60
\\
UniformFace\cite{duan2019uniformface} & MS-Celeb-1M, VGGFace2 & 97.70 
\\
GroupFace\cite{kim2020groupface} & MSCeleb-1M & 97.80 
\\
FSENet\cite{cheng2019face} & Ms-Celeb-1M, VGGFace2 & 97.89 
\\
ArcFace\cite{deng2019arcface} & - & 98.02 
\\
BiometricNet\cite{ali2020biometricnet} & Casia, MS1M-DeepGlint & 98.06 
\\
\hline
\end{tabular}
\end{center}
\end{table*}

\section{Loss Functions \& Activation Functions} 
\label{sec:Loss function and Activation function}

\subsection{Different Loss Functions}%- Awal, Fuad\\
%--fuad\\
The use of loss functions is an essential factor in machine learning-based methods. It helps the machine to learn and predict the results. If the expected result defers from the actual result in the training time, then loss functions try to minimize the difference and help \hl{to} generate a better prediction. In Table \ref{tab:loss_functions}, the classical loss functions and recent loss functions are described shortly.

%--awal

\subsubsection{Contrastive loss} 
Contrastive loss is a distance-based loss function used to compute the distance between the actual output and the predicted output. It provides the pairwise distance between two points through an equation. Contrastive loss can be shown like this:
\begin{equation}
   D_w(X_1,X_2)=||G_w(X_1)-G_w(X_2)||
\end{equation}
Here, we need to optimize the shared parameters w. $G_w(X_1)$ and $G_w(X_2)$ are the two points in the low-dimension space that generated by mapping images $x_1$ and $x_2$.
If $x_1$ and $x_1$ belong to different class then contrastive loss function value will be large. Otherwise, the value will be small.

\subsubsection{Triplet loss} 
Triplet loss\cite{triplet_loss} mainly focuses on both intra-class and inter-class difference. It creates a triplet which consists of baseline $x_b$, positive image $x_p$ and negative image $x_n$.  Mathematically it is defined by: 
\begin{equation}
    TL(x_b, x_p, x_n) = max(||x_b-x_p||^2 -||x_b-x_p||^2+\alpha, 0)
\end{equation}
At first, three face images are needed to be provided where two of them are from the same person, and the third one is from a different person. This loss function's objective is to minimize the distance from the baseline to the positive image ($x_b-x_p$) and maximize the distance from the baseline to the negative image ($x_b-x_n$). The negative image should be away from the positive image by a margin $\alpha$, just like SVM. It is mainly used for face verification tasks. Liu et al.\cite{liu2015targeting} used this loss in their matrix learning step to reduce the features' dimension. FaceNet\cite{facenet} also used this loss in their work.

\subsubsection{N-pair loss}
N-pair loss\cite{n-pair_loss} is more general version of \hl{triplet} loss. It is applied on $N+1$ images. Among them $N-1$ are negative images and one positive image. The N-pair loss with $N+1$ example is defined by:
\begin{equation}
    NP({x,x^+,\{x_i\}_{i=1}^{(N-1)}};f)=log(1+\sum_{i=1}^{N-1}exp(f^T f_i-f^T f^+))
\end{equation}
Here, $f$ is an embedding kernel. The deep neural network defines it. $x^+$ is the only positive example, and $x_1, ..., x_{N-1}$ are negative examples.
When the number of negative examples is one, it works similarly to triplet loss.

\subsubsection{Marginal loss}
By minimizing the intra-class variances and maximizing the inter-class distances of the in-depth features, Deng et al. proposed a loss function to enhance discriminative power called marginal loss\cite{deng2017marginal}. It mainly focuses on the marginal example and tries to minimize the difference between them. The function of marginal loss can be shown like this:
\begin{equation}
    L_m=\frac{1}{m^2-m}\sum_{i,j,i\neq j}^m(\xi-y_{ij} (\theta-||\frac{x_i}{||x_i||}-\frac{x_j}{||x_j||}||_{2}^2))_+
\end{equation}

Here, $x_i$ and $x_j$ are input images. $\theta$ is the threshold margin, and $\xi$ is the error margin for classification. $y_{ij}$ becomes -1 or 1 depending on whether $x_i$ and $x_j$ are in the same class or not. $(u)_+$ indicates that u is positive or zero. For marginal loss, $||x_i-x_j||$ is close to $\theta$ when $x_i$ and $x_j$ are from the same class otherwise is considerably away from $\theta$. Marginal loss can work individually or along with other traditional loss functions like Softmax.

\subsubsection{Ring loss}
Feature normalization through traditional normalization results in a non-convex formulation. To solve this, Zheng et al.\cite{ring_loss} proposed an elegant normalization approach for the deep neural network called Ring loss. Mathematically it can be written as $L_R$ and shown in equation \ref{eq:5}.

\begin{equation}
    L_R=\frac{\lambda}{2m}\sum_{i=1}^m(F(x_i)-R)
\label{eq:5}
\end{equation}

For image $x_i$, $F(x_i)$ is the feature from deep neural network. Here, $m$ is the batch size, and $\lambda$ is the variable weight that has significant impact on the main loss function $L_R$. Moreover, the target norm value \hl{$R$} is also learned. In Ring Loss, Softmax, large-margin Softmax, or SphereFace is used as the primary loss function.\\

%--fuad

\begin{table*}[ht]
  \begin{center}
    \caption{Loss Functions.}
    \label{tab:loss_functions}
    \begin{tabular}{l  l} 
    \hline
      \textbf{Name} & \textbf{ Function}\\
      \hline\hline
      Contrastive Loss & $D_w(X_1,X_2)=||G_w(X_1)-G_w(X_2)||$ \\
      \hline
      Triplet Loss\cite{triplet_loss} & $TL(x_b, x_p, x_n) = max(||x_b-x_p||^2 -||x_b-x_p||^2+\alpha, 0)$\\
      \hline
      N-pair Loss\cite{n-pair_loss} & $NP({x,x^+,\{x_i\}_{i=1}^{(N-1)}}; f)=\log(1+\sum_{i=1}^{N-1}e^{(f^T f_i-f^T f^+)})$ \\
      \hline
      Marginal Loss\cite{deng2017marginal} & $L_m=\frac{1}{m^2-m}\sum_{i,j,i\neq j}^m(\xi-y_{ij} (\theta-||\frac{x_i}{||x_i||}-\frac{x_j}{||x_j||}||_{2}^2))_+$ \\
      \hline
      Ring Loss\cite{ring_loss} & $L_R=\frac{\lambda}{2m}\sum_{i=1}^m(F(x_i)-R)$ \\
      \hline
      COCO Loss\cite{liu2017learning} & $ L^{COCO} = \sum_{i\epsilon\beta}L^{(\,i)\,} = -\sum_{k,i} {t_k}^(i) \log p_k^{(\,i)\,} = -\sum_{i\epsilon\beta} \log p_{l_i}^{(\,i)\,} $ \\
      \hline
      Softmax Loss & $ f(\,s)\,_i = \frac{e^{s_i}}{\sum_j^C e^{s_j}} ; CE = - \sum_i^C {t_i}{\log(\,f(\,s_i)\,)\,} $\\
      \hline
      L-Softmax Loss\cite{liu2016large} & $ L_i = -\log (\,\frac{e^{||W_{y_i}||||x_i||\psi(\,\theta_{y_i})\,}}{e^{||W_{y_i}||||x_i||\psi(\,\theta_{y_i})\,} + \sum_{j\neq y_i} e^{||W_j||||x_i||\cos{\theta_j}}} )\, $\\
      \hline
      Hierarchical SoftMax Loss &$ p(\omega = \omega_0)=\prod_{j=1}^{L(\omega)-1} \sigma([[n(\omega,j+1) = ch(n(\omega,j))]].v^{\prime T}_{n(\omega,j)}h)  $\\
      & $where [[x]] = \big\{_{-1 \quad otherwise}^{1 \quad if \: x \: is \: true}$\\
      \hline
      A-Softmax Loss\cite{liu2017sphereface} & $L_{ang} = \frac{1}{N} -\log (\,\frac{e^{||x_i||\cos{m\theta_{y_i,i}}}} {e^{||x_i||\cos{m\theta_{y_i,i}}}+{}\sum_{j\neq y_i}e^{||x_i||\cos{\theta_{j,i}}} })\,$ \\
      \hline
      AM-Softmax Loss\cite{wang2018additive} & $L_{AMS} = - \frac{1}{n} \sum_{i=1}^n \log{\frac{e^{s.(\cos{(\theta_{y_i})}-m)}}{e^{s.(\cos{(\theta_{y_i})}-m)} + \sum_{j=1,j\neq y_i}^c e^{s.\cos{\theta_j}}}} $\\
      & $ = - \frac{1}{n} \sum_{i=1}^n \log{\frac{e^{s.(W_{y_i}^T f_i -m)}}{e^{s.(W_{y_i}^T f_i-m)} + \sum_{j=1,j\neq y_i}^c e^{s.W_j^T f_i}}} $;\\& where, $m = \cos{\theta_{W_1},P_1} + \cos{\theta_{W_1},P_2}$ \\
      \hline
      L2-Softmax Loss\cite{ranjan2017l2} & minimize $-\frac{1}{M} \sum_{i=1}^M \log \frac{e^{W^T_{y_i}f(x_i)+b_{y_i}}}{\sum_{j=1}^C e^{W^T_j f(x_i)+b_j}}$\\& subject to $||f(x_i)||_2 = \alpha, \forall i=1,2,...M. $\\
      \hline
      ArcFace\cite{deng2019arcface} & $L_3 = \frac{1}{N}\sum_{i=1}^N \log \frac{e^{s(\cos{(\theta_{y_i}+m)})}}{e^{s(\cos{(\theta_{y_i}+m)})} + \sum_{j=1,j\neq y_i}^n e^{s\cos{\theta_j}}} $\\
      \hline
      Noisy Softmax Loss\cite{chen2017noisy} & $L = - \frac{1}{N} \sum_i \log \frac{e^{f_{y_i}-\alpha||W_{y_i}|| ||X_i|| (1-\cos{\theta_{y_i}}) |\varepsilon|}}{\sum_{j\neq y_i} e^{f_j} + e^{f_{y_i}-\alpha||W_{y_i}||||X_i||(1-\cos{\theta_{y_i}})|\varepsilon|}} $\\
      \hline
      CosFace\cite{wang2018cosface} &  $L_lmc = \frac{1}{N} \sum_i -\log \frac{e^{s(\cos{(\theta_{y_i},i)}-m)}}{e^{s(\cos{(\theta_{y_i},i)}-m)} + \sum_j \neq y_i e^{s\cos(\theta_j,i)}}, $\\
      & subject to, $W = \frac{W*}{||W*||}, x = \frac{x*}{||x*||}, \cos(\theta_j,i) = W_j^Tx_i $\\
      \hline
      Fair Loss\cite{liu2019fair} & $L = -\frac{1}{N}\sum_{i=1}^N \log\frac{P*_{y_i}(m_i(t),x_i)} {P*_{y_i}(m_i(t),x_i) + \sum_{j=1,j\neq y_i}^n P_j(x_i)}$;where $x_i\epsilon \mathbb{R}^d  $\\
      \hline
      CurricularFace\cite{huang2020curricularface} & $L = - \log\frac{e^{s\cos({\theta_{y_i}+m})}} {e^{s\cos({\theta_{y_i}+m})}+\sum_{j=1,j\neq y_i}^n e^{sN(t^{(k)},\cos{\theta_j})}}, $ \\
      & where, $N(t,\cos{\theta_j}) = \{_{\cos{\theta_j}(t+\cos{\theta_j}),     T(\cos\theta_{y_i})-\cos\theta_j \geq 0}^{\cos\theta_j, T(\cos\theta_{y_i})-\cos\theta_j < 0} $\\
      & and $T(\cos\theta_{y_i}) = \cos(\theta_{y_i}+m) $\\
      \hline 
    \end{tabular}
  \end{center}
\end{table*}

\subsubsection{Congenerous Cosine Loss (COCO Loss)} 
Yu et al.\cite{liu2017learning} proposed this loss by minimizing the cosine distance between samples. It reduces the complexity and normalizes the inputs. It also enlarges the distinction between inter-class and decreases the variation between inner-class.

\subsubsection{Softmax Loss}
\begin{figure}[ht]
    \centering
    \includegraphics[width=80mm,scale=1.0]{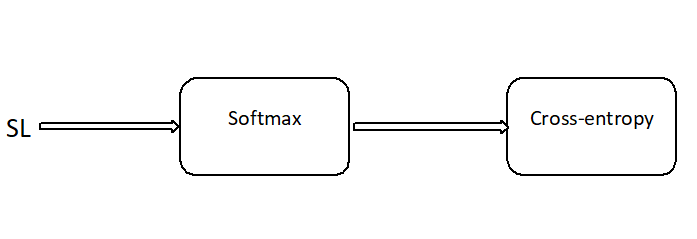}
    \caption{Softmax Diagram.}
    \label{fig:sm_diagram}
\end{figure}
The Softmax or Softargmax function is a generalization of logistic function to multiple dimensions \cite{wikipedia2021}. It is commonly used in deep learning and neural networks. It is a combination of Softmax activation and cross-entropy loss that outputs the probability for every class, and later these will be summed up, shown in Figure \ref{fig:sm_diagram}.

\subsubsection{Large-Margin Softmax Loss (L-Softmax)}

\begin{figure}[htbp]
    \centering
    \includegraphics[width=80mm,scale=1.0]{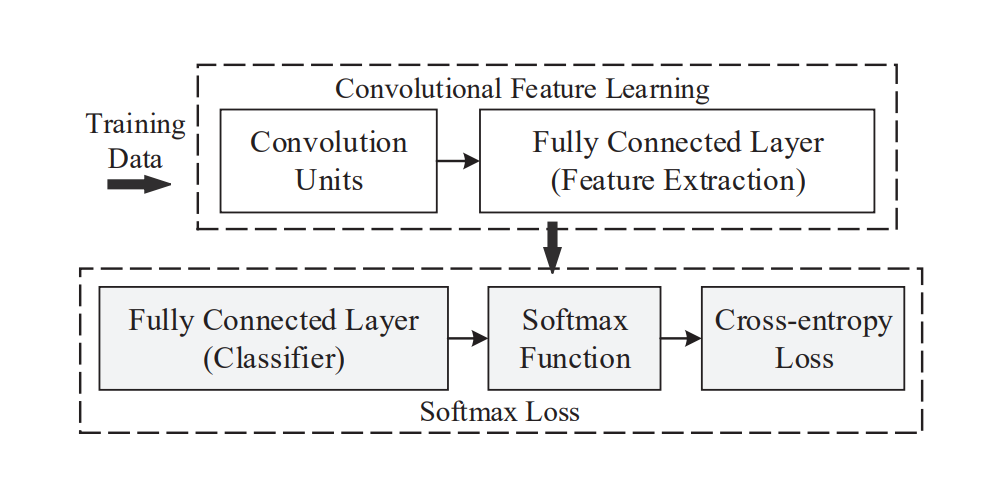}
    \caption{L-Softmax Diagram\cite{liu2016large}.}
    \label{fig:lsm_diagram}
\end{figure}
It is a modified softmax loss function that works with the distances between classes in CNN, proposed by Weiyang et al.\cite{liu2016large}. It tries to maximize the distance between different classes and minimizes the distance between the same classes. As a result, intra-class compactness and inter-class separability boost the performance of recognition and detection tasks. It can also avoid overfitting. Figure \ref{fig:lsm_diagram} shows the workflow of L-Softmax.

\subsubsection{Hierarchical Softmax}
A faster and alternative loss function of Softmax is Hierarchical Softmax. The time complexity of Softmax is $O(n)$, where it can be done by Hierarchical Softmax in just $O(log n)$ time. In computation, it uses a multi-layer binary tree where each class is in the trees' leaf node, and each edge contains a probability value. The probability is calculated with the product of values on each edge from the root to that node from that tree. The main advantage of hierarchical Softmax is that it works faster than the Softmax function. Wang et al.\cite{H-softmax-FR} used hierarchical Softmax with ABASNet in their multi-face recognition method.

\subsubsection{Angular Softmax Loss (A-Softmax)}
Weiyang et al.\cite{liu2017sphereface} proposed a model called SphereFace where they used a new loss that incorporates the angular margin. They changed the decision boundary of softmax loss to $||W_1||=||W_2||=1 $ and $b_1=b_2=0$, which are weight and bias. So, the decision boundary becomes $||x||(cos(\theta_1)-cos(\theta_2))=0$. Then, the decision boundary only depends on $\theta_1$ and $\theta_2$. It directly optimizes the angles, enabling CNNs to learn angularly distributed features. They used an integer $(m \geqslant 1)$ to quantitatively control the decision boundary and used it with $\theta_1$ and $\theta_2$ respectively in two classes, where $m$ controls the size of the angular margin.

\subsubsection{Additive Margin Softmax Loss (AM-Softmax)}
Wang et al.\cite{wang2018additive} proposed AM-Softmax for deep face verification. It is instinctively appealing and more efficient than the existing loss in margin-based work. It normalizes the weight and bias like A-Softmax. It uses an entirely new hyperparameter $s$, which measures the cosine value. The decision boundary is also adjusted according to the loss. 

\subsubsection{L2-Softmax Loss}
For solving the performance gap of similarity score between positive pairs and negative pairs, Rajeev et al.\cite{ranjan2017l2} proposed a new loss L2-constrained Softmax Loss. It has an L2-constraint in feature descriptors which constricts the features to be on a fixed radius hypersphere by keeping the L2-norm constant. Forcing it to stay in a fixed radius minimizes the cosine similarity between the negative and positive pairs.

\subsubsection{Additive Angular Margin Loss (ArcFace)}
Jiankang et al.\cite{deng2019arcface} proposed a new loss function called Additive Angular Margin Loss (ArcFace). It can obtain highly discriminative features for FR and stabilize the training process. Moreover, it also has a clear geometric interpretation due to its exact correspondence to geodesic distance on a hypersphere. The dot product between the DCNN feature and the last fully connected layer is equal to the cosine distance after feature and weight normalization. The arc-cosine function is used to determine the angle between the current feature and the target. The authors fixed the bias as zero, transformed the logit, fixed the individual weight by l2 normalization, fixed the embedding feature, and rescaled it.

\subsubsection{Noisy Softmax Loss}
Binghui et al.\cite{chen2017noisy} tried to solve softmax loss's early saturation behaviour by proposing noisy softmax loss. It migrates the early saturation problem by injecting annealed noise for every iteration and also brings continuous gradient propagation, which dramatically encourages the SGD solver. 

\subsubsection{CosFace: Large Margin Cosine Loss (LMCL)}
Wang et al.\cite{wang2018cosface} proposed a novel loss function, namely Large Margin Cosine Loss (LMCL). They applied L2 normalizing in both features and weight vectors and reformulated the softmax loss as a cosine loss. Then, a cosine margin term $m$ was introduced to maximize the decision margin in the angular space, dubbed Large Margin Cosine Loss (LMCL). The model, which is trained with LMCL, is named as CosFace. Authors contributed in 3 ways: (1) proposed a novel loss function called LMCL, (2) provided theoretical analysis for LMCL. (3) advanced the state-of-the-art performance over most of the famous face databases.

\subsubsection{Fair Loss}
For solving the imbalance problem, Bingyu et al.\cite{liu2019fair}  proposed a new margin-aware reinforcement learning-based loss using Deep Q-Learning. They explored the adaptive boundaries between classes and proposed to balance the additive margins between various classes. They imitated all changes in additive margins for classes in the training process and collected influence on the training model. They concluded a strategy of adaptive margin by using Deep Q-learning.

\subsubsection{CurricularFace: Adaptive Curriculum Learning Loss}
Yuge et al.\cite{huang2020curricularface} proposed a credible method named CurricularFace using Adaptive Curricular Learning. CurricularFace solves convergence issues of features. It mainly addresses easy samples in preliminary stages and complex samples in later stages. Firstly, the curriculum construction is adaptive; the samples are randomly selected in each mini-batch. The curriculum is established adaptively via mining the hard online, which shows the diversity in samples with different importance. Secondly, the priority of complex samples is adaptive. The misclassified samples in mini-batch are chosen as complex samples and weighted by adjusting the modulation coefficients of cosine similarities between the sample and the non-ground class vectors.

\subsection{Different Activation Functions}% -- Awal

The activation function defines output depending on a given single input or a set of inputs in a node. It detects how much that node will contribute to the next node or nodes. The activation function can be both linear or non-linear. For deep learning, non-linear activation functions are highly preferable. Those can give a highly accurate output than the linear activation functions. There are many types of non-linear activation function. Table \ref{tab:Activation Function} shows a brief overview of commonly used activation functions.

\begin{table*}[ht]
  \begin{center}
    \caption{Activation Functions.}
    \label{tab:Activation Function}
    \begin{tabular}{l l l} 
    \hline
      \textbf{Name} & \textbf{ Function} & \textbf{Output Range}\\
      \hline\hline
      Sigmoid & $f(x)=\frac{1}{1+e^{-x}}$  & (0, 1)\\
      
      Tanh& $f(x)=\frac{e^{-x}-e^{x}}{e^{-x}+e^{x}}$ & (-1, 1)\\
      
      Softmax& $f(x)= \frac{e^{x_i}}{\sum_{i=0}^N e^{x_i}}$ & (0, 1)\\
      
      ReLU\cite{relu}& $f(x)=\big\{_{0\quad x\leq0}^{x\quad x > 0}$ & $[0, \infty)$\\
      
      Softplus\cite{softplus} & $f(x)=ln(1+e^x)$ & $[0, \infty)$\\
      
      Leaky RelU\cite{LReLU}& $f(x)=\big\{_{0.01x\quad x\leq0}^{x\quad x > 0}$  & $(-\infty, \infty]$\\
      
      Parametric ReLU\cite{PReLU}&  $f(x)=\big\{_{ax\quad x\leq0}^{x\quad x > 0}$  & $(-\infty, \infty]$\\
      
      ELU\cite{ELU}&  $f(x)=\big\{_{a(e^x-1)\quad x\leq0}^{x\quad x > 0}$  & - \\
      
      Swish\cite{swish}& $f(x)=\frac{x}{1+e^{-\beta x}}$ & -\\
      
      Maxout\cite{maxout}& $max(w_1^{T}x+b_1, w_2^(T)x+b_2)$ &$[-\infty, \infty]$\\
      \hline
    \end{tabular}
  \end{center}
\end{table*}

\subsubsection{Sigmoid}
Sigmoid is one of the most popular probability functions.  There are many types of sigmoid activation functions; one of them is the logistic activation function. It takes all possible values as input and provides output in the range of 0-1. Its output is an "S"-shaped curve, also called a sigmoid curve. It is mainly used in the last layer to predict the output. It can show the probability of a new data point of being that class. The main problem with sigmoid is that it can not handle the vanishing gradient problem. It can only predict two classes; it is not possible to classify in multi-class with sigmoid activation. Some other sigmoid activations are used in face recognition are Adjustable Generalized Sigmoid, Sigmoidal selector.

\subsubsection{Tanh} 
Hyperbolic tangent Activation Function, also known as Tanh activation function, is another sigmoid type activation function. Its output graph is also 'S'-shaped curve ranging between -1 to 1. It is mainly used in feed-forward neural networks. Being zero means it can handle both positive and negative values easily. It works better than the sigmoid activation function in almost all situations. Like sigmoid, it also cannot handle the vanishing gradient problem and can not classify in multi-class.

\subsubsection{Softmax}
Softmax is another popular activation function for DCNN, DNN, machine learning models. It is primarily used in the last layer for multi-class classification. It converts the output as a vector of probabilities of that data is in each class. The sum of the possibilities is one. It can take positive, negative, or zero, all possible values. It provides one output for every possible class in a normalized form in the range of 0 to 1. It solves the main problem of sigmoid and tanh activation function. It can classify into multi-class. Almost all CNN or DCNN based face identification \hl{researches}, including \cite{wu2017recursive}, \cite{DDRL_FV_FI} used softmax in their last layer.

\subsubsection{ReLU}
Rectified Linear activation function or ReLU\cite{relu} is one of the most popular and sometimes default activation functions for many DCNN based models. It is a piecewise linear activation function that takes all possible values as input and output only when the values are positive and sets all negative values as zero.
Its output range is 0 to infinity. Solving the vanishing gradient problem is not possible for Sigmoid and Tanh activation function, but it can be solved with ReLU. ReLU is faster than most of the other activation functions.
ReLU also faces some problems. Its centre is not at zero and has no maximum limit. It sometimes comes to a state where the neuron becomes inactive and stuck there, especially in the first few layers. Also, no backpropagation can take the neuron out of it. It is called the dying ReLU problem. All popular DCNN based models for face recognition models use ReLU in their internal convolution layers. A  smoother version of ReLU is Softplus\cite{softplus}. Some other methods, including \cite{encrypted} use softplus as an activation function for their face recognition methods.

\subsubsection{Leaky ReLU}
Leaky Rectified Linear Activation, also known as Leaky ReLU or L-ReLU\cite{LReLU}, is also a piecewise activation function that works with the same idea as ReLU. The only difference between ReLU and Leaky ReLU is when the input value is negative. Instead of setting zero like ReLU when the value is negative, Leaky ReLU multiplies the value with a small number $a$ (generally .01). So the negative portion gets a value but very small. It is an attempt to solve the dying ReLU problem. However, linearity is the main problem of leaky ReLU. So it can not be used in complicated classification tasks. Also, it is hardly possible to find out the perfect multiplayer value $a$.

\subsubsection{Parametric ReLU}
Parametric Rectified \hl{Linear} Activation or Parametric ReLU or P-ReLU\cite{PReLU} is another version of Leaky ReLU. However, unlike Leaky ReLU, P-ReLU takes the slope of the negative portion as parameter $a$. The neural network finds it through gradient descent. It solves the problem of a predefined multiplier from Leaky ReLU. Nevertheless, it creates a new problem; it can act differently in a different situation.
%--- Challenges table
\begin{table*}[t]
  \begin{center}
    \caption{Overview of Still Image-based Face Recognition.}
    \label{tab:sifr}
    \begin{tabular}{p{1.5in}p{1.5in}p{1.5in}p{1.5in}}
    \hline
     \textbf{Algorithm} & \textbf{Method} & \textbf{Dataset} & \textbf{Accuracy (\%)} \\
      \hline\hline
      ReST\cite{wu2017recursive} & CNN & LFW & 99.03\\
      &     & YTF & 95.40\\
      
      DR-GAN\cite{tran2017disentangled} & GAN & IJB-A & 94.7\\
      & & CFP & 97.84 ± 0.79\\
      
      Xiao et al.\cite{chal3} & GAN & CMU MultiPIE & 99.2\\
      
      Peng et al.\cite{peng2017reconstruction} & DNN & CMU MultiPIE & 90.5\\
      & & 300WLP & 98.0\\
     
      HPDA\cite{wang2020hierarchical} & CNN & LFW & 99.80\\
      & &CACD -VS & 99.55\\
      
      PW-GAN\cite{chal4} & GAN & LFW & 98.38\\
      
      LF-CNN\cite{wen2016latent} & DCNN & LFW & 99.5\\
      & & CACD & 98.5\\
      
      Li et al.\cite{chal9}& CNN & MORPH & 93.6\\
      & & CACD-VS & 91.1\\
      
      OE-CNNs\cite{caf_db} & CNN & Morph Album 2 & 98.67\\
      & &CACD-VS & 99.5\\
      & &LFW & 99.47\\
      
      Wang et al.\cite{chal11}& DAL & MORPH Album 2 & 98.93\\
      & & CACD -VS & 99.40\\
      
      Choi et al.\cite{chal12} & DCNN & CMU Multi-PIE & 96.24\\
      
      Li et al.\cite{chal9} & BLAN & Ms-celeb-1m  & 99.8\\
      & & Dataset 2\cite{chal37} & 82.7\\
      & & FAM \cite{chal38} & 97.0\\
      
      RGMS\cite{chal16} & DNN & ORL & 98.75\\
      && YALE & 98.67\\ 
      && AR dataset & 92.30\\
      
      NNAODL\cite{chal14} & DL & LFW & 93.1\\
      && AR database & 99.0\\
      && ExYaleB & 99.7\\
      \hline

    \end{tabular}
  \end{center}
\end{table*}
%---------

\subsubsection{Exponential Linear Unit (ELU)}
Exponential Linear Unit was also known as ELU\cite{ELU}. It is also a piecewise activation function. It provides the same value as input, when the value is positive. However, when the input value is negative, its output is $\exp{(x)}-1$ multiplied by a constant value. The constant value is generally 0.1 or 0.3. As a result, it does not suffer from vanishing and exploding gradients problem. As it does not stick at zero on a negative value, so does not suffer from dying neuron as ReLU. Moreover, the most significant advantage is that it provides higher accuracy, and training timing is faster than ReLU in the neural network. Another type of ELU that also used in face recognition is Parametric ELU.

\subsubsection{Swish}
Swish\cite{swish} also provides a ReLU like output, where the input value is highly negative. The difference is that it does not change certainly at zero. From zero, it bends towards a negative value depending on a variable and creates a smooth curve. For positive values, it provides a positive output. In the Deep Neural Network test, swish always performs better than ReLU with every batch size. It is used in face recognition in \cite{swish_FD}.

\subsubsection{Maxout}
Maxout\cite{maxout} is a simple piecewise activation function that provides the maximum of the input. It is a generalization of ReLU and Leaky ReLU activation functions. It takes advantage of the ReLU unit but does not have drawbacks. The main problem with maxout is that it doubles the number of computations in each neuron. As a result, it is much slower than ReLU. FaceNet\cite{facenet} model used maxout activation in their fully connected layer for face recognition.

\section{Challenges in Face Recognition Using Deep Learning}
\label{sec:Challenges in Face Recognition Using Deep Learning}
%-- Fuad\\
Many challenges can be seen when we use face recognition in real-life scenarios. Those challenges keep us from getting the perfect accuracy. Deep learning methods try to solve the drawbacks and significantly improve accuracy. In recent years researchers focused on solving the challenges. We can notice several challenges in still image-based face recognition (SIFR), video-based face recognition (VFR), heterogeneous face recognition (HFR), etc.

\subsection{Still image-based face recognition}
We can see more than half of the published papers on face recognition are on solving the SIFR challenges in the past. The problems are solved using CNN, AE, GAN, RL, etc. The researchers focus on solving pose variations problems, cross-age, illumination changes, facial makeup and expression variations. Table \ref{tab:sifr} shows the summary of still image based face recognition methods.

\subsubsection{Pose-invariant}
Nowadays, CNN-based models of face recognition have two-step pipeline: face detection and face recognition. In the ReST\cite{wu2017recursive} paper, the authors discussed the problems of this two-step pipeline. Sometimes the alignment step transforms all faces into the same, and this causes geometrical information loss. We can see diversity when it comes to different poses, illumination, etc. However, in the two-step pipeline system, we lose this, which is essential for differentiation objects. To solve this problem, they design a novel Recursive Spatial Transformer module for CNN. It optimizes face alignment and recognition jointly in one network in an end-to-end system. The recursive structure has three parts: Convolutional layers, Localization network and Spatial Transformation layer. Here, the whole face is divided into hierarchical layers of regions, and each region is equipped with a ReST. It tries to handle large face variations and non-rigid transformations.
\par 
In DR-GAN\cite{tran2017disentangled}, the author used Generative Adversarial Network for pose variations. They used an encoder-decoder structure-based Disentangled Representation Learning. In \cite{chal3} Xiao et al. proposed a Geometric Structure Preserving-based GAN for multi-pose face frontalization. Here the perception loss compels the generator part to adjust the face image with the same input image. In the discriminator part, the self-attention block is used to preserve the geometry structure of a face. Sufang et al. \cite{chal4} worked with large pose and photo-realistic frontal view synthesis variations in a generic manner and proposed a Pose-Weighted Generative Adversarial Network (PW-GAN). To solve problems like not being photo-realistic and losing ID information, they frontalized the face image through the 3D face and gave more attention to large poses, and they refined the pose code in the loss function. \par 
Xiangyu et al.\cite{chal5} proposed a method HPEN to recover the frontal face pose variation,  which can recover the canonical-view of images using a 3D morphable model that automatically generates the face from frontal pose and expression. They created a 3D landmark from a 2D image using 3DMM (morphable method). Then, they used the mesh technique for the invisible position, and the invisible region was filled with the Trend fitting and Detail fitting method. However, this method's main drawback is that it performs poorly when it comes to occluded images, and there is no clearance for large databases or real-time feedback.
Peng et al.\cite{peng2017reconstruction} tried to reconstruct from pose-invariant based images in their DNN based FR model. They reconstruct the 3D shape from a near-frontal face to generate new face images. They generate a non-frontal view from the frontal image and search the identity of the large embedded feature of identity and pose-variance. They also developed a feature reconstruction metric to learn the identity. 
\par 
Wang et al.\cite{wang2020hierarchical} proposed a pyramid diverse attention (PDA) to learn multiscale diverse local representation automatically and adaptively. They claimed this model reduces problems like pose variations or large expressions or similar local patches. They developed the model HPDA by fusing HBP and PDA. In HPDA, it can describe diverse local patches at various scales adaptively and automatically from varying hierarchical layers. Here, it guides multiple local branches in each pyramid scale to focus on diverse regions instead of face mark landing and a hierarchical bilinear pooling is combined. It also uses different cross-layer bilinear modules to integrate both high and low levels. This model has four parts: stem CNN, local CNN, global CNN, and classification. They use HSNet-61 in the background mainly. They also fused SENet and HSNet model. They used their own proposed divergence loss in diverse learning to guide multiple local branches to learn diverse attention masks. The diverse learning encourages each local branch to learn different attention masks by increasing their distances. 
\hl{Ding et al. \mbox{\cite{ding2016comprehensive}} briefly discussed pose-invariant face recognition in their survey paper. The authors quoted the problems of PIFR as well as discussed possible future directions of Face Recognition tasks.}

%------

\subsubsection{Age-invariant}
Age is always an essential factor in Face Recognition.\hl{ We know that with the change of age, face changes.} So, recognizing faces becomes more complicated when the test sample is aged. The researchers tried to solve this problem by experimenting with many deep learning models.
Following it, Yandong et al.\cite{wen2016latent} proposed a deep CNN based age invariant face recognition named LF-CNN for deep face features. They extracted the age-invariant deep features from convolutional features by a carefully designed fully connected layer, termed as (LF-FC) layer. They developed a latent variable model, called latent identity analysis (LIA), to separate the variations caused by the aging process from the identity-related components in convolutional features. This model has two components: convolutional unit for feature learning and latent factor fully connected layer for age-invariant deep feature learning. They also used PReLU and max-pooling for enhancing robustness.
\par 
Li et al.\cite{chal9} proposed a novel distance metric optimization technique that integrates feature extraction and the application of distance metrics and interaction between them using DCNN. It learns feature representation with an end-to-end decision function. They collected images from different age instances. Then they enlarged the differences between the unmatched pairs by reducing variations among matched pairs simultaneously. They used the mini-batch SGD algorithm to update the parameters, the top fully connected layer of the distance matrix, and the image features from the bottom layer.
%video face table
%
\begin{table*}[ht]
  \begin{center}
    \caption{\hl{Overview of Video Face Recognition.}}
    \label{tab:vfr}
    \begin{tabular}{l l l l}
    \hline
      \textbf{Algorithm} & \textbf{Method} & \textbf{Dataset} & \textbf{Accuracy (\%)} \\
      \hline\hline
      NAN \cite{yang2017neural} & DCNN & IJB-A & 0.986 ± 0.003\\
      && YTF & 95.72 ± 0.64\\
      && Celebrity-1000 & 90.44\\
      
      FBA \cite{face-body-video} & DNN & JANUS CS3 & 85.3\\
      
      ADRL\cite{adrl} & RL & YTF & 96.52 ± 0.54 \\
      && YTC & 97.82 ± 0.51\\
      
      Wang et al\cite{chal28} & DCNN & Chinese citizen & 98.92±0.005 (imbalanced)\\
      && face image dataset[28] & 94.36 ± 0.01 (balanced)\\
      
      Liu et al \cite{chal30} & RL & IJB-A & 97.3±1.1\\
      && YTF & 96.01±0.48\\
      &&Celebrity-1000 & 91.37\\
      \hline
    \end{tabular}
  \end{center}
\end{table*}
\par
The intra-class discrepancy has always been a problem in face recognition, especially in age-invariant problems. Wang et al.\cite{caf_db} proposed a novel Orthogonal Embedding CNNs (OE-CNNs) which decomposed the deep face features into two orthogonal components. It represents age-related and identity-related features. They used A-Softmax loss because different identities are discriminated by different angles and decomposed in spherical coordinates with radial coordinate and angular coordinates. The decomposed features improve performance. 
In reducing discrepancy on AIFR, Wang et al.\cite{chal11} also proposed a novel algorithm. They tried to remove age-related components from features mixed with identity and age information. They factorized a new mixed face feature into two non-correlated elements: identity-dependent and age-dependent. They proposed the Decorrelated Adversarial Learning (DAL) algorithm, and a Canonical Mapping Module (CMM) was introduced, which found the maximum correlation of the paired features. The model learns the decomposed attributes of age and identity. To ensure the correct information, it simultaneously supervised the identity-dependent attribute and the age-dependent attribute. The proposed model has an extension of CCA, the Batch Canonical Correlation Analysis (BCCA). This method significantly increases the state-of-the-art (SOTA) on AIFR datasets.

Besides age and pose-invariant challenges in still image-based face recognition, we can see many challenges, such as facial makeup, illumination changes, partial face, facial expression, etc. Recently many researchers have started work on these challenges. Choi et al.\cite{chal12} used a DCNN model for eliminating illumination effects and maximizing discriminative power. Zhao et al.\cite{chal13} used a modified local binary pattern histogram (LBPH) for solving illumination diversification, expression variation and attitude deflection. Du et al.\cite{chal14} proposed a framework for illumination changes and occlusion in face recognition named Nuclear Norm based Adapted Occlusion Dictionary Learning (NNAODL). They used a two-dimensional structure and dictionary learning (DL) in their framework. Li et al.\cite{chal15} proposed a bi-level adversarial network (BLAN) for makeup problems in FR. To overcome posture, illumination and expression problems, Mona et al.\cite{chal16} proposed a novel approach called Relative Gradient Magnitude Strength (RGMS) for feature extraction. This method is based on Deep Neural Networks (DNNs).

\subsection{Video-based face recognition}
\hl{Video-based FR (VFR) is difficult in comparison with still image-based face recognition.} When it comes to VFR, various problems come forward. \hl{Most of the videos usually come from mobile, which causes large pose variations, occlusions, out-of-focus blur, motion blur, etc.} On the other hand, surveillance cameras, CCTV cameras cause cross-domain problems, and low-quality problems, etc. Researchers tried to partially solve pose-variations and occlusion in SIFR using the embedding technique \cite{chal17}, \cite{liu2015targeting}, but in VFR, the techniques are not extended. Some methods which are mainly proposed for SIFR but also work quite well for VFR, for example, DeepFace\cite{chal19}, DeepID2\cite{chal20}, FaceNet\cite{facenet}, VGGFace\cite{parkhi2015deep}, C-FAN\cite{chal23} etc. In C-FAN, Sixue et al. trained the model using CNN for SIFR and learned the quality value-added to an aggregation module. It performs well in VFR as it aggregates deep feature vectors in a single vector for face in the video. Table \ref{tab:vfr} shows the overview of video based face recognition methods.

Jiaolong et al.\cite{yang2017neural} proposed a Neural Aggregation Net-work (NAN) for VFR. As input, it takes a set of face images or face video and produces a compact, fixed-dimension feature representation. They used DCNN for feature embedding, and for face verification, Siamese neural aggregation network and minimized average contrastive loss is used. On the other hand, a fully connected layer followed by a softmax and classification loss is used for identification. Kim et al.\cite{face-body-video} proposed a novel approach, face and body association (FBA) in VFR. They used a retrained YOLO detector in face detection and a single DNN with ResNet-50 as backbone architecture in verification. For a video frame, they extract 18 key points in the 2D joints of the skeleton person. The data association stage has a scoring function, greedy data association, tracklet initialization and termination, tracklet filtering, and parameter settings like sub-stages. However, it treats the face and upper body as similar. Recently, there are some works on VFR using deep reinforcement learning such as ADRL\cite{adrl}, automatic face ageing\cite{duong2019automatic} etc. For the real-time video, Wang et al.\cite{chal28}, and Grundstrom et al.\cite{chal29} proposed DCNN based models. Liu et al.\cite{chal30} proposed a dependency aware attention control (DAC) model, which used a reinforcement learning-based sequential attention decision of image embedding.

% Image Datasets table
\begin{table*}[ht]
  \begin{center}
    \caption{\hl{Image Datasets for Face Recognition.}}
    \label{tab:image_database}
    \begin{tabular}{l l l l} 
    \hline
      \textbf{Name} & \textbf{ Number of} & \textbf{Total Images} & \textbf{Description}\\
      \textbf{} & \textbf{Individuals} & \textbf{} & \textbf{}\\
      \hline
      \hline
      AgeDB\cite{agedb} & 570 & 16,516 & Manually collected images of age range from 1 to 101. All of \\ &  & & those images are taken from a uncontrolled environment with \\ & & & different pose, lighting, and noise.\\
      
      Large Age-Gap (LAG)\cite{large_age_gap_db} &  1,010 & 3,828 & A dataset with images of people with large age difference.\\
    
      CAF\cite{caf_db} & 4,668 & 313,986 & It is a \hl{noise} free dataset containing images collected from \\ & & & Google. It also contains some images of  Asian individuals.\\
      
      CAFR\cite{cafr_db} & 25k & 1,446,500 & All images are  annotated with identity, gender, age,  and race. \\ & & &The age range is from 1 to 99 and divided into 7 age phases.\\
      
      CPLFW\cite{CPLFW_db} & 5,749 & 11,652 & Pose difference and the number of images are more balanced \\ & & & than LFW.\\
     
      Trillion-Pairs\cite{trillion_pair} &  5.7k & 274k & Mainly used for testing. It is divided into two parts: ELFW, \\ & & & DELFW. \\
      
      IMDb-Face\cite{IMDb} & 59K & 1.7M & Clean dataset collected from movie screenshots and posters.\\
      
      MS1M-DeepGlint\cite{trillion_pair} & 86,876 & 3,923,399 & Large scale aligned face for training.\\
      
      Asian-DeepGlint\cite{trillion_pair} & 93,979 & 2,830,146 &Large scale aligned face for training.\\
      
      GANFaces-500k\cite{GANFaces-500k_and_5M_db} & 10k & 500k  & Synthetic dataset mainly used for training.\\
      
      GANFaces-5M\cite{GANFaces-500k_and_5M_db} & 10k & 5,000k & Synthetic dataset mainly used for training.\\
      
      LFW\cite{LFW} & 5,749 & 13,233 & Images with variation in pose,  ethnicity, lighting, age,\\ & & & expression,   background, gender, clothing, hairstyles, camera,\\ & & & quality, color saturation, and other parameters.\\
      
      CelebFaces\cite{sun2013hybrid} & 5, 436 & 87, 628 & Images of celebrities collected from web used for training.\\
      
      CASIA-WebFace\cite{casia-webface-dataset} & 10,575 &  494,414 & Contains images of celebrities who were born between\\ & & & 1940 to 2014 and mainly used for training.\\
      
      VGG Face\cite{parkhi2015deep} & 2,622 & 2.6M & A large dataset of publicly available images for training. \\
      
      VGG Face2\cite{cao2018vggface2} & 9,131 & 3.31M & A large dataset for training with a large variations in pose, \\ & & & age, illumination, ethnicity and profession.\\
      
      MegaFace\cite{megaface} & 690k & 1M & Used as a image gallery.\\
      
      MS-Celeb-1M\cite{MS-Celeb-1M} & 100k & 10M & Images of celebrities, mainly for training set.\\
      
      RMFRD\cite{RMFRD_SMFRD_db} & 525 &  95k & Two types of images for same persons: wearing mask and \\ & & &  without wearing mask.\\
      
      SMFRD\cite{RMFRD_SMFRD_db} &  10k & 500k & Use of a software to automatically create faces with mask on\\ & & &  popular face datasets.\\
      
      KomNET\cite{komnet} & - & 39.6k & Those images are collected from 3 different \hl{sources}: phone\\ & & & camera, digital camera and social media  without \\ & & & considering lighting,  mustache, beard, background, haircut, \\ & & &expression, and head covered glasses.  \\
       \hline
    \end{tabular}
  \end{center}
\end{table*}
%-----

\subsection{Heterogeneous face recognition}
Besides VFR and SIFR, Heterogeneous Face Recognition remains a challenging problem as cross-modality has limited training samples as well as complicated generation procedure of face images. Cao et al.\cite{chal31} proposed a GAN-based asymmetric joint learning (AJL) process, which transforms the cross-modality variance. Wu et al.\cite{chal32} proposed a CNN-based coupled deep learning (CDL) method to seek a shared feature space. In this method, heterogeneous images are treated as homogeneous images. Di et al.\cite{chal33} proposed a hybrid model using GAN and CNN, which focused on extracting images from the visible range for synthesizing and took thermal images as input. He et al.\cite{chal34} also proposed CFC, a GAN-based model for solving heterogeneous face synthesis problems. Jos et al. \cite{chal35} proposed to extend the DL breakthrough for VIS face recognition to the NIR spectrum without retraining the underlying deep models that see only VIS faces. It has two core integrants, cross-spectral hallucination, and low-rank embedding. Cross-spectral hallucination produces VIS faces from NIR images through a DL approach. Low-rank embedding restores a low-rank structure for the deep features of faces across both the NIR and VIS spectrum. \hl{Ouyang et al.\mbox{\cite{ouyang2016survey}} discussed briefly in their survey paper on heterogeneous FR. Here, they quoted NIR-based faces, sketch-based faces,  3D faces, low-resolution images, etc. They also discussed their observation on the paper and some future directions, for example, computing time, datasets, alignment, technical methodologies, training volume, etc.
}
%--video dataset description for face recognition%

\begin{table*}[ht]
 \begin{center}
 \caption{\hl{List of Video Datasets for Face Recognition.}}
    \label{tab:videoDatasetTable}
\begin{tabular}{p{1in}p{1in}p{1in}p{3in}}
\hline
 \textbf{Name}  & \textbf{Total Videos} &  \textbf{Number of Individuals} &\textbf{Description}\\
\hline\hline
IARPA Janus Benchmark A (IJB-A)\cite{yang2017neural} & 2,042 & 500 & The videos contain pose variation.
\\
IARPA Janus Benchmark-B (IJB-B)\cite{8014821} & 7,011 & 1,845 & Here, approximately 4 videos/subjects are available and  average of 30 frames for each subject. This dataset contains more geographical distribution.
\\
IARPA Janus Benchmark-C (IJB-C)\cite{8411217} & 11,779 & 3,531&This dataset have been created after removing the celebrity divergence. More pose variations are also included.
\\
YouTube Face (YTF)\cite{yang2017neural} & 3,425 & 1,595 & The lengths of videos vary from 48 to 6,070 frames. The average video length is 181.3 frames.\\ 

YouTube Celebrities (YTC)\cite{li2018face} &1,910 &47 & All the entity has been converted into MPEG4 format. The videos has 25fps rate.
\\
Celebrity-1000\cite{yang2017neural} & 159,726 & 1,000 &The
videos contain celebrity images with 15 frames per second.\\

FaceSurv\cite{gupta2019facesurv} & 460 & 252 & Benchmark face detection and face recognition dataset with different spectra and resolutions.
\\
UMDFaces\cite{bansal2017dosanddonts}& 22,075  & 3,107 &3,735,476 video frames have annotated from 22,075 videos. This dataset also figured out pose (pitch, roll, yaw), twenty-one key-points location and generated the gender information.
\\
PaSC \cite{6712704} & 2,802 &265 & The videos have been collected based on many types of variations, such as location (inside and outside of buildings), pose, distance (both near and far from camera) and one video camera control with five held video cameras.
\\
\hl{VDMFP\mbox{\cite{scheirer2016report}}} & 958 & 297 & The dataset contains two types of videos: walking and conversation. The videos were collected to avoid various constraints such as illumination and pose.
\\
\hline 

\end{tabular}
\end{center}
\end{table*}

%heterogeneous table
\begin{table*}[ht]
  \begin{center}
    \caption{\hl{Heterogeneous Face Datasets for Face Recognition.}}
    \label{tab:HFD}
    \begin{tabular}{l l l l} 
    \hline
      \textbf{Name} & \textbf{ Number of} & \textbf{Total Images} & \textbf{Description}\\
      \textbf{} & \textbf{Individuals} & \textbf{or Videos} & \textbf{}\\
      \hline
      \hline
      
      CUHK VISNIR\cite{gong2017heterogeneous} & 2,800 & 5,600 & Each person has two images one is optical another is near\\
       & & & infrared photo.\\
      
      NIR-PF \cite{he2016multiscale}& 276 & 5,300 & Contains 16-20 NRI images of each subject with different\\
       & & &view, lighting condition, distance and scale. \\
      
      Polarimetric thermal \cite{hu2016polarimetric}& 60 & - & Contains LWIR and VIS data from 3 different distances.\\
       
      MGDB \cite{ouyang2016forgetmenot} & 100 & 400 &  Four face sketch of each subject; 3 of them are drawn after\\
       & & &3 different time duration from viewing the mugshot by the\\
       & & &artist and 1 by hearing face description from the eyewitness.\\
       
      e-PRIP \cite{mittal2017composite} & 123 & 123 & Extends PRIP \cite{han2012matching} database by adding sketch images for\\
      & & & each subject.\\
       
      UoM-SGFS \cite{galea2016large}& 300 & 600 & Contains software generated sketches.\\
       
      Extended UoM-SGFS \cite{galea2017matching} & 600 & 1,200 & Contains software generated colored sketches. \\
       
      UHDB31 \cite{wu2016rendering} & 77 & 1,617 & Captured images form 21 viewpoints for each subject.\\
      
      LS3DFace\cite{gilani2018learning} &  1,853 &  31,860  & Contains 3D images.\\
       
      Lock3DFace \cite{zhang2016lock3dface}& 509 & 5,711 & Contains  video clips of RGBD face with different occlusion\\
      & & & pose, time lapse and facial expression.\\
       
      RGB-D DB \cite{cui2018improving} & 747 & 845k &  With a few illumination change and continuous pose \\
      & & &variations in RGB-D \hl{format}.\\
       
      NJU-ID \cite{huo2016ensemble} & 256 & 13,056 & Fifty one images per subject in different \hl{resolution}.\\
      
      ID-Selfie-A \cite{shi2018docface}& - & 20,000 & Ten thousand pair of selfies are taken from a stationary \\
      & & & camera and ID photos from chips.\\
      
      ID-Selfie-B \cite{shi2018docface} & 547 & 10,844 & Individual ID card images and variable number of selfies \\
      & & &for each individual.\\
       \hline
    \end{tabular}
  \end{center}
\end{table*}
%-----
\section{Face Datasets}
\label{sec:Face Datasets}
%-- Awal, Delowar
\subsection{Image Datasets}
%--Awal
\hl{Face recognition is a complex task in the real world scenario.} To do it perfectly large and correctly labelled training dataset is required. Collecting face images and label them properly is a time-consuming task. \hl{There are a lot of publicly available datasets those can be used for this purpose.} Earlier datasets were small in size, less than hundred identities. As time passes, more researchers and companies have come into this field. They are investing their time and money, so the size of the datasets is getting large. Some of the publicly available datasets have already crossed a few million face images\cite{IMDb}, \cite{GANFaces-500k_and_5M_db}. Nowadays, most of the images to create a new dataset are collected from different social media or websites\cite{komnet}. The main problem for face recognition from the images is that most of the features from the face change with the change \hl{of} the pose or age. Pointing out this problem, some researchers added images of different pose and age limit\cite{LFW}, \cite{CPLFW_db}. Some datasets also contain synthetic face images to increase the number of images in their collection, for instance, GANFaces500k\cite{GANFaces-500k_and_5M_db}. After covid-19 breaks out, face recognition with face masks getting researchers' attention. Some datasets of people with masks and without masks like RMFRD\cite{RMFRD_SMFRD_db}, SMFRD\cite{RMFRD_SMFRD_db} are already publicly available. Table \ref{tab:image_database} shows some of the recent available datasets.

\subsection{Video Datasets}
%--Delowar \\
\hl{Face recognition from video data is a great issue in this era}. So the video-based FR machine learning algorithm has become more popular nowadays. Many videos data have been generated through YouTube, Facebook, Instagram, and other social media. However, for this process, more videos data need to be trained through machines, so that the model can achieve excellent performance. It is obvious that large-scale datasets are needed to show better accuracy for face recognition from video data. 
To improve the Video-based FR task, some excellent work has been done and the researchers collected video data that helped to enhance the accuracy of the system.\hl{ It is also noted that five groups from world renowned institutions work on Point-and-Shoot Challenge (PaSC) video data to evaluate the accuracy of PittPatt algorithm\mbox{\cite{beveridge2015report}}.}
Here, Table \ref{tab:videoDatasetTable} shows different video face datasets (e.g. IJB-A, YTF, IJB-B, YTC, and IJB-C etc.) for face recognition and explains their properties.

\subsection{Heterogeneous Face Datasets}

%--Awal

Heterogeneous face recognition is a challenging but important. It is used in different types of applications like security and law enforcement. HFR is a problem of recognizing face from images of nontraditional sources of light such as Near Infrared (NRI), Sketch, or 3D images. Images of NRI datasets are taken under infrared instead of visible light. CUHK VIS-NIR \cite{gong2017heterogeneous}, NIR-PF\cite{he2016multiscale} contains image under infrared. Sketch images are human art of other human's face. MGDB\cite{ouyang2016forgetmenot}, e-PRIP\cite{mittal2017composite} are recent sketch face image datasets for face recognition. On the other hand, LS3DFace \cite{gilani2018learning} and Lock3DFace \cite{zhang2016lock3dface} databases contain 3D faces. \hl{Although HFR is getting popular and some datasets are available, most of the datasets are small in size}. Table \ref{tab:HFD} shows a brief overview of the recently available HFR datasets.

\section{Future Trends}
\label{sec:Future Trends}
\subsection{Dataset size and training time}
\hl{DNN based face recognition has come a long way. Currently, the state-of-the-art networks can take millions of images to train and manage hundreds of millions of parameters to generate output. Some of those methods showed incredible results on testing datasets. But still, DNN has a long way to go. Most of the methods took a long time and large dataset to train. Researchers can search the way to develop methods which can be trained with small dataset and take short time. Bio-inspired methods can be a great help for them. }
\subsection{COVID-19}
\hl{As the COVID-19 has broken out in recent years, some security measures have been taken to save the human species from this pandemic. Wearing masks is one of them, and for this reason, traditional face recognition methods are mostly useless in this unexpected situation. So, researchers should pay attention and search for new ways to detect faces or persons. Some researchers \mbox{\cite{RMFRD_SMFRD_db}} have already started their work. But more should come and join them. The use of infrared cameras can be a good solution. Moreover, researchers can think of this type of scenario for FR. 
}
\subsection{Face Recognition in Infrared Faces}
\hl{Whenever obstacles come between face and camera, regular face photos are not adequate for FR. To overcome this issue, the researchers can focus on Infrared (IR) face images nowadays. IR images provide a multi-dimensional imaging system. The multi-dimensional imaging system is used to get more accurate results in unfavorable conditions like object illumination, expression changes, facial disguises, and dark environments\mbox{\cite{arya2015future}}. So, the researchers can improve algorithms which are focused on IR-based FR.}
\subsection{COST FUNCTION}
\hl{
In recent times, researchers have tried to improve the cost functions. They can try to merge existing loss functions like Softmax loss and Centre loss\mbox{\cite{zhang2021facial}, \cite{center-loss}}. They can also try to use various cost functions in different layers like yang et al.\mbox{\cite{yang2020fan}}. Mainly, the researchers need to find more efficient cost functions to decrease the computational time.
}

\section{Conclusion}
\label{sec:Conclusion}
%--Delowar
\hl{Our paper has demonstrated the recent advances of Deep learning-based face recognition systems that are mainly focused on algorithms}, architecture, loss functions, activation functions, datasets, and varied types of occlusion such as pose-invariant, illusion, expression of face, age, and variations of ethnicity etc. Most of the Face Recognition systems has been built using Deep learning and the architecture may be changed according to the dataset variations and performance improvement issues. Deep learning architecture has shown excellent performance in the face recognition systems in recent decades.
Different types of datasets like still image-based, heterogeneous face image-based, video-based, and occlusion-based datasets are shown in our paper as summarized forms. Our paper found that LFR, IJB, YTF and  Ms-celeb-1M have shown near perfect performance in various FR tasks.
Occlusion based challenges still appear in the FR task. This situation hampers the performance of the FR systems. More datasets and novel algorithms may reduce the occlusion based problems. Despite some limitations and challenges of the face recognition tasks, these systems are improved significantly in recent years.

\bibliographystyle{IEEEtran}
\bibliography{IEEEabrv, reference}

\begin{IEEEbiography}[{\includegraphics[width=1in,height=1.25in,clip,keepaspectratio]{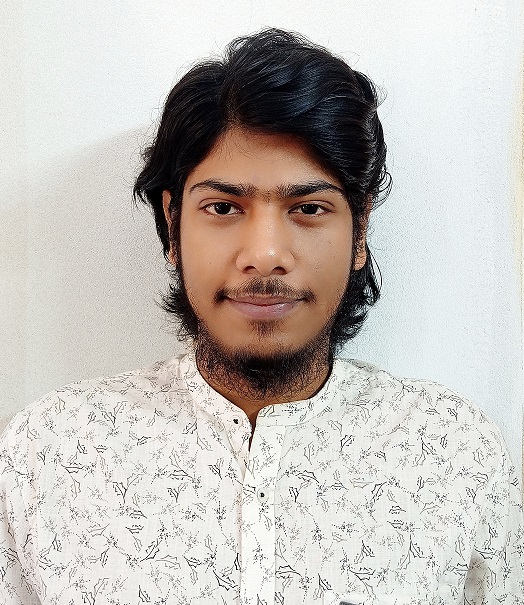}}]{Md. Tahmid Hasan Fuad} was born in Rajshahi, Bangladesh. He is currently pursuing a B.Sc. degree in Computer Science and Engineering (CSE) with the Khulna University of Engineering \& Technology (KUET), Khulna, Bangladesh. His research interests include Image Processing, Computer Vision, Artificial Intelligence, Machine Learning, and Deep Learning. He has already worked on some exciting Android Development and Machine Learning based mini-projects. He has also done some mini-projects using C++, Python, Java etc.
\end{IEEEbiography}
\vskip -1\baselineskip plus -1fil
\begin{IEEEbiography}[{\includegraphics[width=1in,height=1.25in,clip,keepaspectratio]{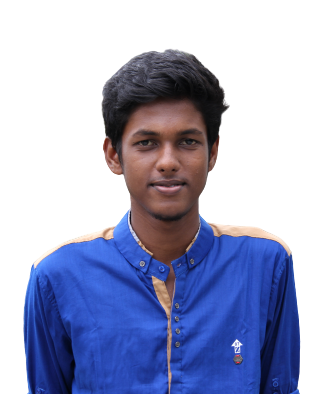}}]{Awal Ahmed Fime}
was born in Jashore,
Bangladesh. He is currently pursuing the B.Sc.
degree in Computer Science and Engineering (CSE)
 with the Khulna University of Engineering \&
Technology (KUET), Khulna, Bangladesh. His
research interests include Computer vision, Artificial Intelligence, Signal processing,
Machine learning, and Deep learning.
He has already worked on some Web and Mobile Application using ASP.NET, CSS, JavaScript, Android
throughout his study using latest technology.
\end{IEEEbiography}
\vskip -1\baselineskip plus -1fil
\begin{IEEEbiography}[{\includegraphics[width=1in,height=1.25in,clip,keepaspectratio]{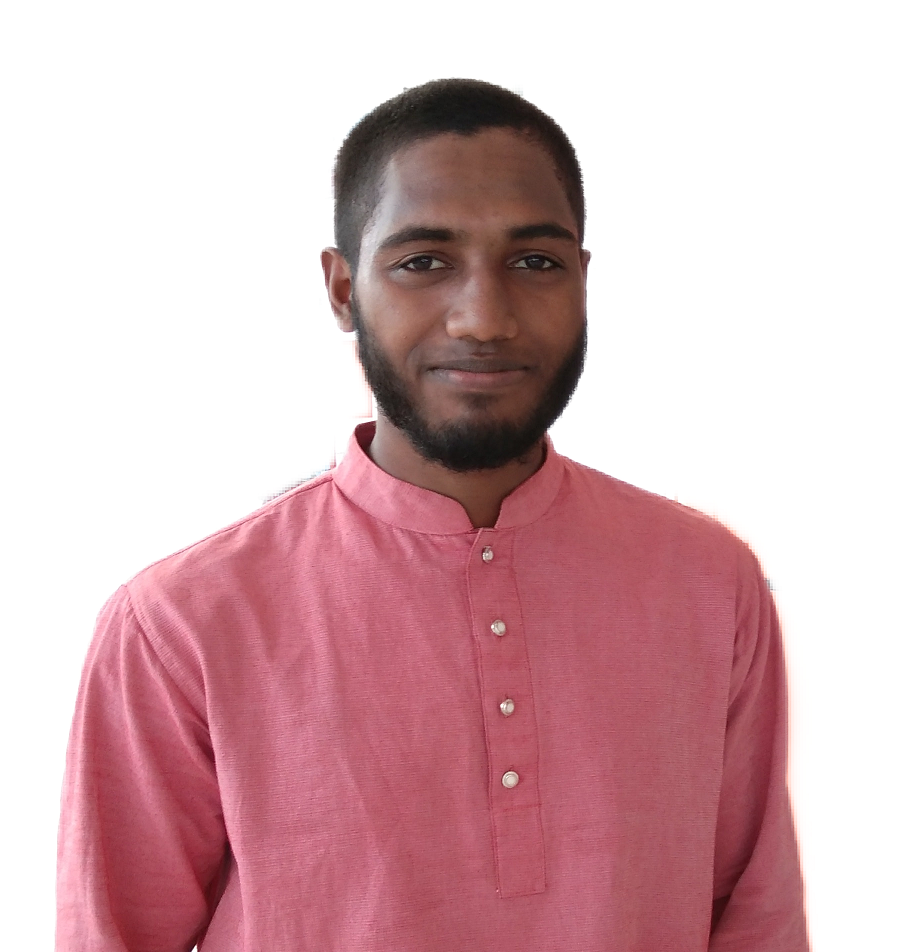}}]{Delowar Sikder}
was born in Patuakhali,
Bangladesh. He is currently pursuing the B.Sc.
degree in Computer Science and Engineering (CSE)
 with the Khulna University of Engineering \&
Technology (KUET), Khulna, Bangladesh. His
research interests include Computer vision, Artificial Intelligence,
Machine learning, and Deep learning.
He has also keen interested to Automated system design, Web and Mobile Application Development. 
He has already worked on some interesting projects
throughout his study using latest technology.
\end{IEEEbiography}
\vskip -1\baselineskip plus -1fil
\begin{IEEEbiography}[{\includegraphics[width=1in,height=1.25in,clip,keepaspectratio]{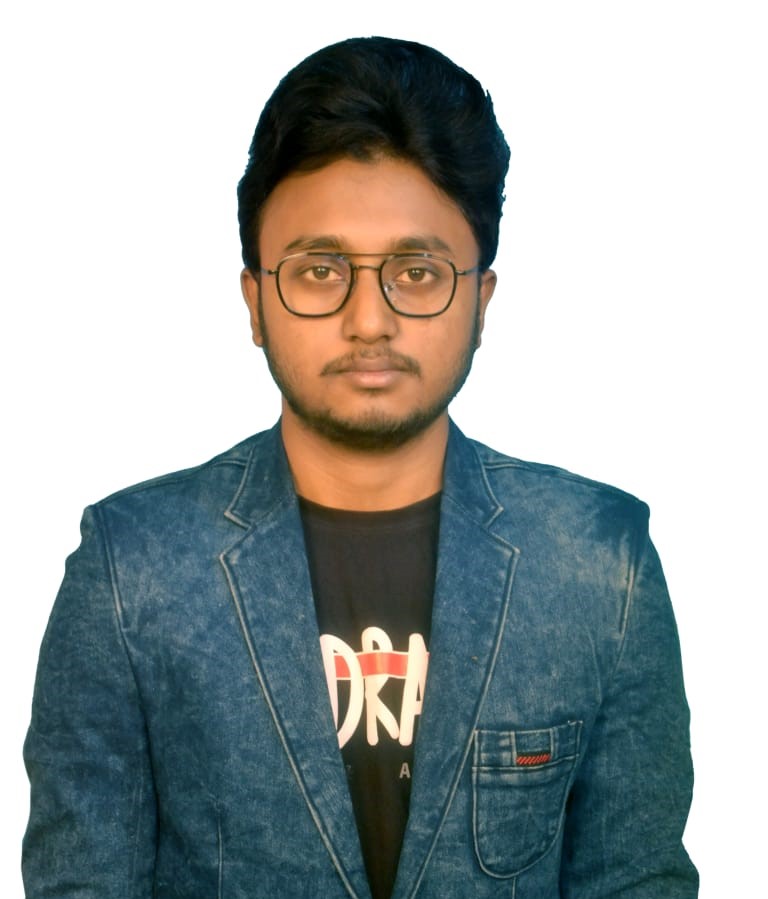}}]{MD.AKIL RAIHAN IFTEE}
was born in  
Joypurhat,  Bangladesh. He is currently pursuing a B.Sc. degree in Computer Science and Engineering (CSE) with the Khulna University of Engineering \& Technology (KUET), Khulna, Bangladesh. His research interests include  Deep  Learning, Data Science, Artificial Intelligence, Machine Learning, and Natural Language Processing. He has already developed some projects using C, C++, Python, Java, HTML, SQL, etc. He is a regular participant in machine learning and data science competitions on an online platform such as Kaggle, Hacker-Earth, etc.
\end{IEEEbiography}

\vskip -1\baselineskip plus -1fil
\begin{IEEEbiography}[{\includegraphics[width=1in,height=1.25in,clip,keepaspectratio]{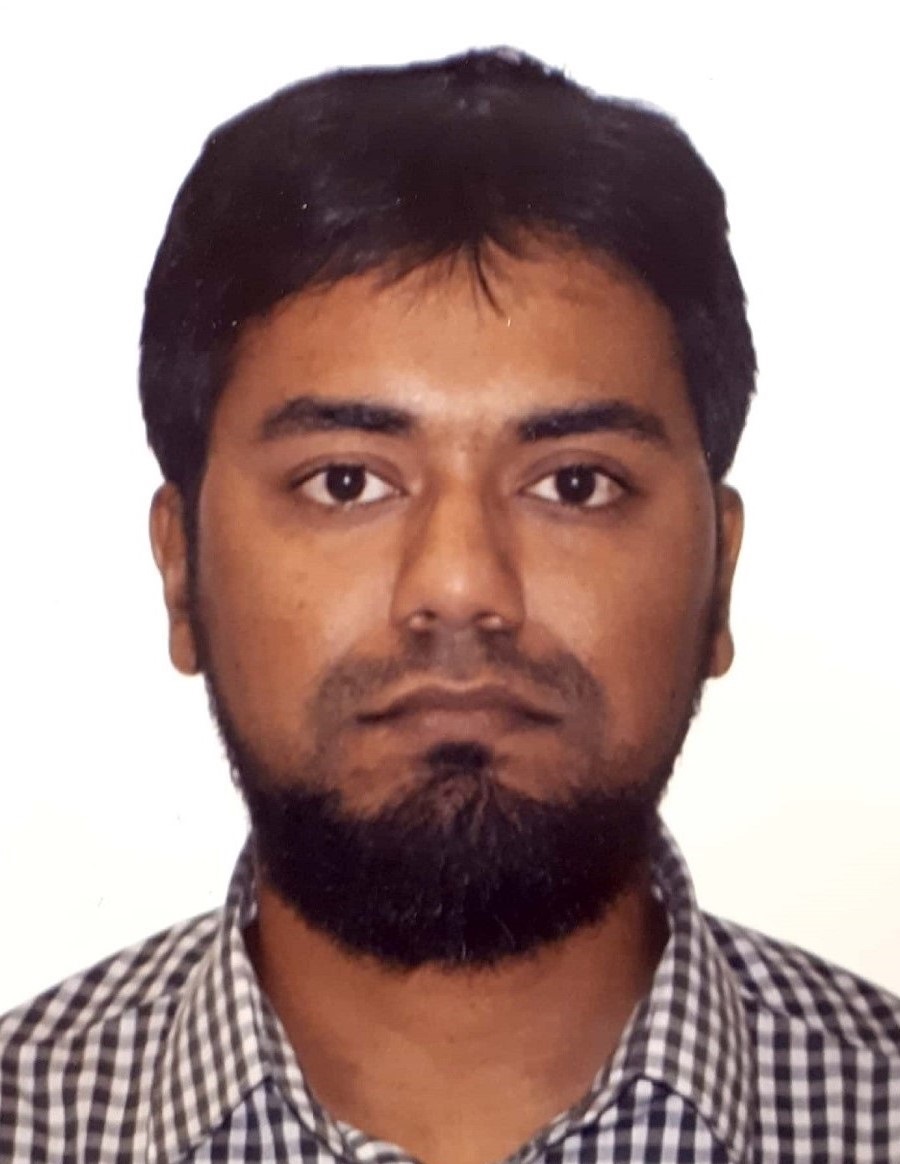}}]{Jakaria Rabbi} received a master's degree in Computing Science from University of Alberta, Edmonton, Canada. He is currently working as an Assistant Professor at the department of Computer Science and Engineering (CSE), Khulna University of Engineering \& Technology (KUET), Khulna, Bangladesh. His research interests include  Machine Learning, Deep Learning, Computer Vision Artificial Intelligence, Data Science and Remote Sensing. He has authored and coauthored several articles in peer-reviewed Remote Sensing journal and IEEE conferences.
\end{IEEEbiography}

\vskip -1\baselineskip plus -1fil
\begin{IEEEbiography}[{\includegraphics[width=1in,height=1.25in,clip,keepaspectratio]{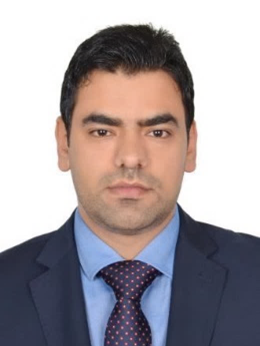}}]{MABROOK S. AL-RAKHAMI}
 (M’20–SM’16) received a master’s degree in information systems from King Saud University, Riyadh, Saudi Arabia, where he is currently pursuing a Ph.D. degree with the Information Systems Department, College of Computer and Information Sciences. He has worked as a lecturer and taught many courses, such as programming languages in computer and information science, King Saud University, Muzahimiyah Branch. He has authored several articles in peer-reviewed IEEE/ACM/Springer/Wiley journals and conferences. His research interests include edge intelligence, social networks, cloud computing, Internet of things, big data and health informatics.
\end{IEEEbiography}

\vskip -1\baselineskip plus -1fil
\begin{IEEEbiography}[{\includegraphics[width=1in,height=1.25in,clip,keepaspectratio]{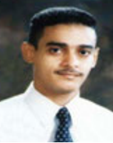}}]{ABDUL GUMAEI} received the Ph.D. degree in computer science from King Saud University, in 2019. He worked as a Lecturer and taught many courses, such as programming languages at the Department of Computer Science, Taiz University, Yemen. He is currently an Assistant Professor with the College of Computer and Information Sciences, King Saud University, Riyadh, Saudi Arabia. He has authored and coauthored more than 30 journal and conference papers in well-reputed international journals. He received a patent from the United States Patent and Trademark Office (USPTO) in 2013. His research interests include software engineering, image processing, computer vision, machine learning, networks, and the Internet of Things (IoT).
\end{IEEEbiography}
\vskip -1\baselineskip plus -1fil

\begin{IEEEbiography}[{\includegraphics[width=1in,height=1.25in,clip,keepaspectratio]{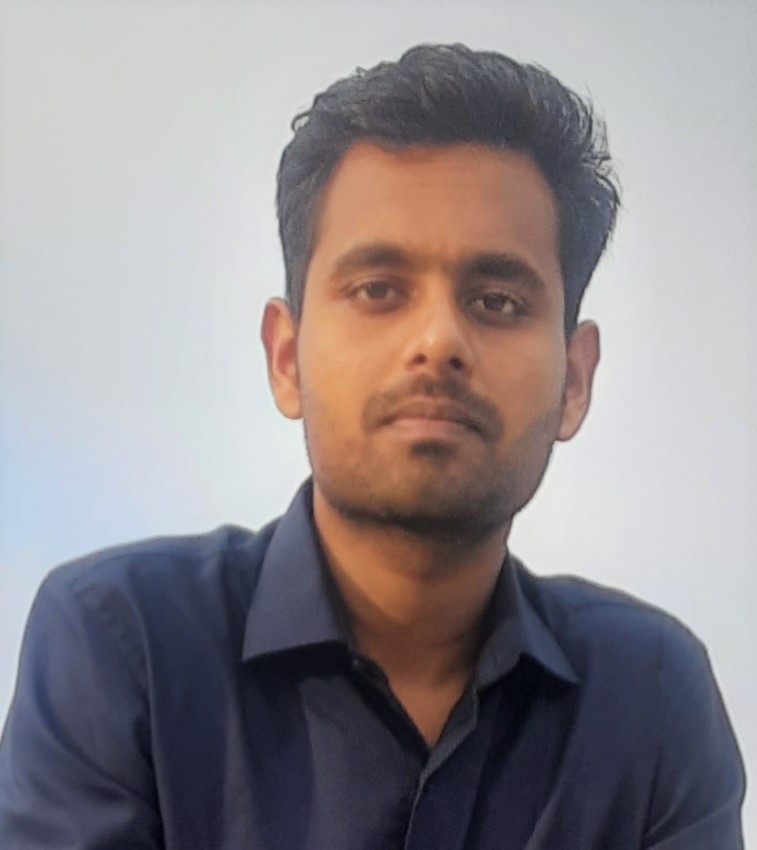}}]{OVISHAKE SEN} was born in Thakurgaon, Bangladesh. Currently, he is pursuing a B.Sc. degree in computer science and engineering (CSE) with the Khulna University of Engineering \& Technology (KUET), Khulna, Bangladesh. He has developed some exciting projects using C, C++, Python, Java, HTML, CSS, ASP.net, SQL, Android, and iOS. His research interests include natural language processing, computer vision, speech processing, machine learning, deep learning, competitive programming, and data science.
\end{IEEEbiography}
\vskip -1\baselineskip plus -1fil
\begin{IEEEbiography}[{\includegraphics[width=1in,height=1.25in,clip,keepaspectratio]{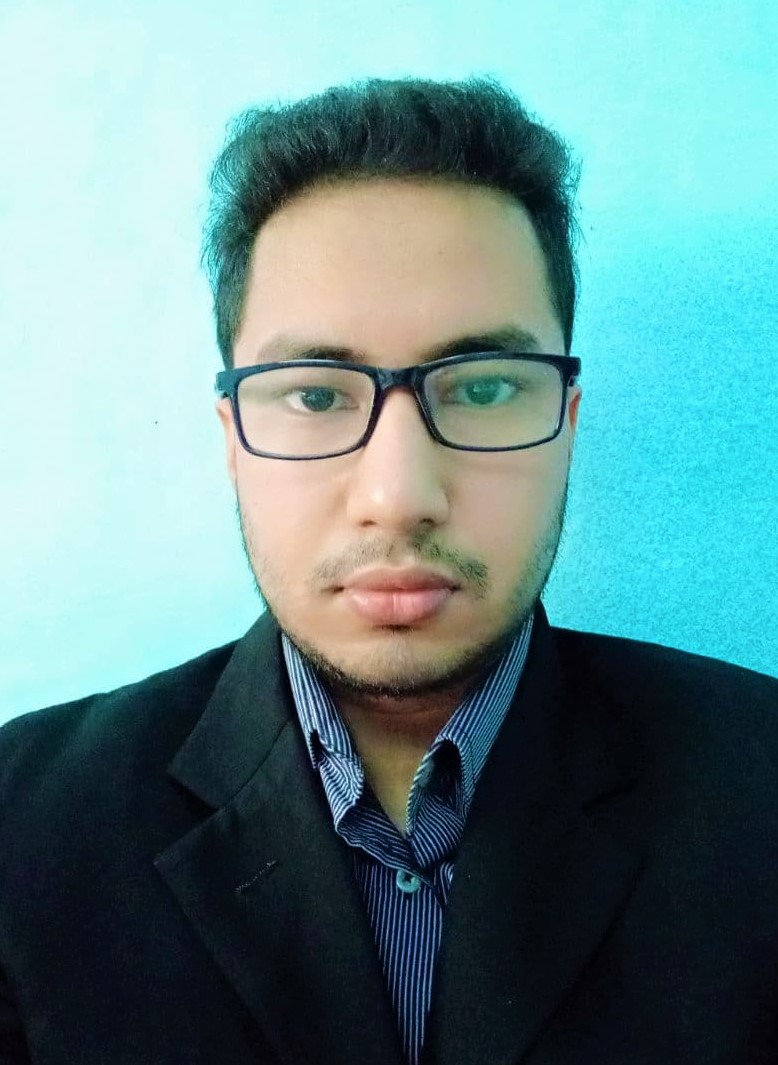}}]{Mohtasim Fuad} was born in Chattrogram, Bangladesh. He is currently pursuing a B.Sc. degree in computer science and engineering (CSE) with the Khulna University of Engineering \& Technology (KUET), Khulna, Bangladesh. His research interests include computer vision, natural language processing, data science, machine learning and deep learning. He is currently working on deep learning projects.
\end{IEEEbiography}

\vskip -1\baselineskip plus -1fil
\begin{IEEEbiography}[{\includegraphics[width=1in,height=1.25in,clip,keepaspectratio]{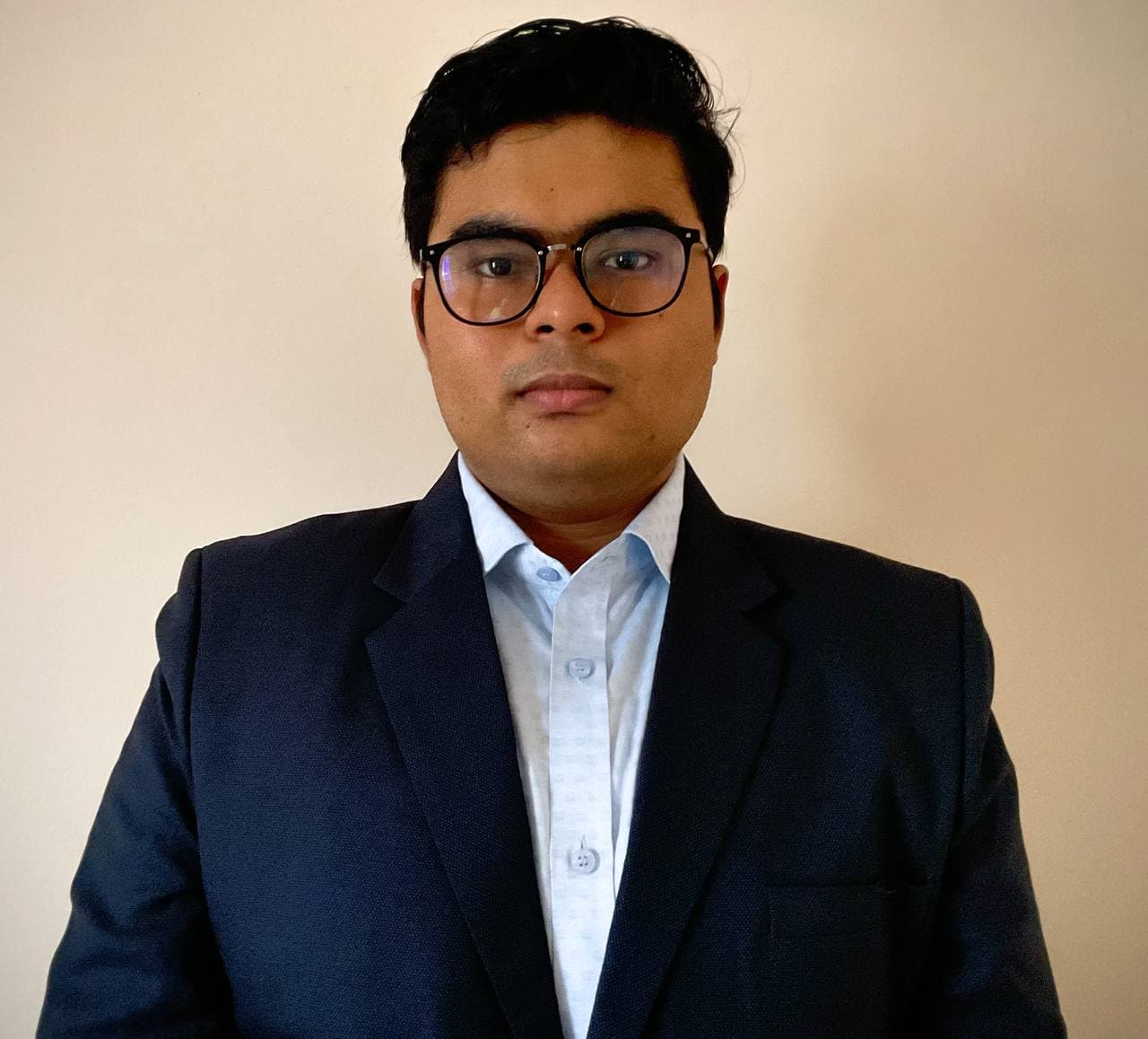}}]{MD. NAZRUL ISLAM} was born in Chandpur, Bangladesh. He is currently pursuing a B.Sc. degree in computer science and engineering (CSE) with the Khulna University of Engineering \& Technology (KUET), Khulna, Bangladesh. He has worked on some exciting projects and some collaborative works throughout his study. He has worked sincerely at one of the data science projects of OneBlood Blood Centers in a  concerted effort with success. His research interests include machine learning, arm architecture, data science, deep learning, computer vision, RISC architecture, and natural language processing.
\end{IEEEbiography}

%\begin{IEEEbiography}[{\includegraphics[width=1in,height=1.25in,clip,keepaspectratio]{a2.png}}]{Second B. Author} 
%\end{IEEEbiography}

%\begin{IEEEbiography}[{\includegraphics[width=1in,height=1.25in,clip,keepaspectratio]{a3.png}}]{Third C. Author, Jr.} (M'87) 

%\end{IEEEbiography}

\EOD

\end{document}